\documentclass[10pt,twocolumn,letterpaper]{article}

\usepackage[accsupp]{axessibility}  
\usepackage{packages/iccv}
\usepackage{times}
\usepackage{epsfig}
\usepackage{graphicx}
\usepackage{amsmath}
\usepackage{amssymb}
\usepackage{authblk}
\usepackage{float}
\usepackage{enumitem}
\usepackage{pdfpages}
\usepackage[table,dvipsnames]{xcolor}

\usepackage[pagebackref=true,breaklinks=true,colorlinks,bookmarks=false]{hyperref}

\iccvfinalcopy 

\def\iccvPaperID{1355} 

\ificcvfinal\fi

\begin{document}
\title{Monocular, One-stage, Regression of Multiple 3D People\vspace{-8mm}}

\author{Yu Sun$^1$\thanks{This work was done when Yu Sun was an intern at JD AI Research. }\quad
Qian Bao$^2$ \quad
Wu Liu$^{2}$\thanks{Corresponding author.} \quad
Yili Fu$^{1\dagger}$\quad
Michael J. Black$^3$ \quad
Tao Mei$^2$\\\vspace{-2.5mm}
$^1$Harbin Institute of Technology \quad $^2$JD AI Research \\
$^3$Max Planck Institute for Intelligent Systems, T\"ubingen, Germany\\
{\tt\small \texttt{yusun@stu.hit.edu.cn, baoqian@jd.com, liuwu@live.cn, meylfu@hit.edu.cn}}\\\vspace{-1mm}
{\tt\small\texttt{black@tuebingen.mpg.de, tmei@live.com}}
}

\maketitle
\ificcvfinal\thispagestyle{empty}\fi

\begin{abstract}
This paper focuses on the regression of multiple 3D people from a single RGB image. Existing approaches predominantly follow a multi-stage pipeline that first detects people in bounding boxes and then independently regresses their 3D body meshes. In contrast, we propose to Regress all meshes in a One-stage fashion for Multiple 3D People (termed ROMP). The approach is conceptually simple, bounding box-free, and able to learn a per-pixel representation in an end-to-end manner. Our method simultaneously predicts a Body Center heatmap and a Mesh Parameter map, which can jointly describe the 3D body mesh on the pixel level. Through a body-center-guided sampling process, the body mesh parameters of all people in the image are easily extracted from the Mesh Parameter map. Equipped with such a fine-grained representation, our one-stage framework is free of the complex multi-stage process and more robust to occlusion. Compared with state-of-the-art methods, ROMP achieves superior performance on the challenging multi-person benchmarks, including 3DPW and CMU Panoptic. Experiments on crowded/occluded datasets demonstrate the robustness under various types of occlusion. The released code\footnote{\url{https://github.com/Arthur151/ROMP}} is the first real-time implementation of monocular multi-person 3D mesh regression.

\end{abstract}
\vspace{-3mm}
\section{Introduction}~\label{sec:intro}
\vspace{-5mm}

\begin{figure}[t]
	\centering
	\includegraphics[width=0.98\columnwidth]{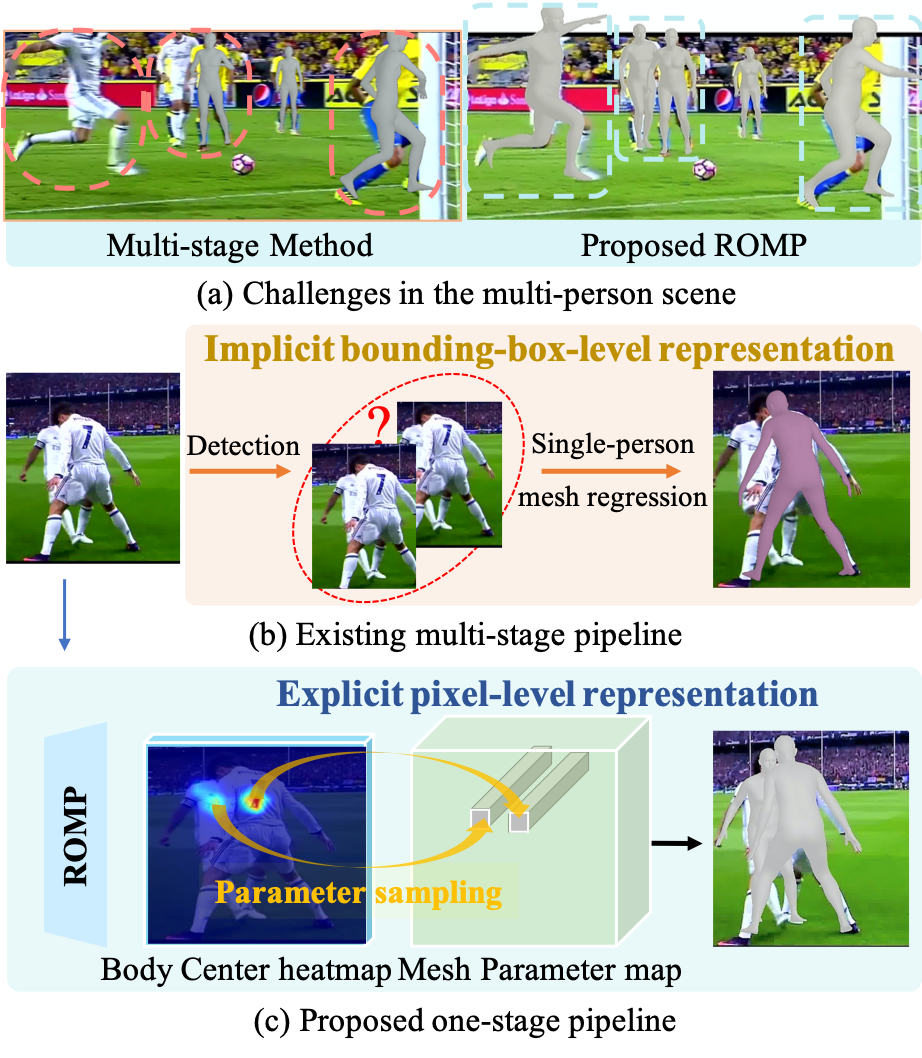} 
	\caption{
    Given a challenging multi-person image like (a), the state-of-the-art approaches, e.g., VIBE~\cite{kocabas2020vibe} (left), fail to deal with truncation, scene occlusion, and person-person occlusion. 
 The reason lies in the multi-stage design (b), where the bounding-box-level features are often implicit, ambiguous, and inseparable in multi-person cases.
 We propose to regress all meshes in one single stage for multiple 3D people.
Specifically, we develop an explicit pixel-level representation (c) for fine-grained one-stage estimation that increases robustness to truncation and occlusion while significantly reducing computational complexity.
}
	\label{fig:motivation}
\end{figure}

Recently, great progress has been made in monocular 3D human pose and shape estimation, particularly in images with a single person~\cite{hmr,kocabas2020vibe,kolotouros2019spin,surreal,Xu_2019_ICCV}.
However, for more general scenes with multiple people, it is crucial to deal with truncation by the image frame, person-person occlusion, and environmental occlusion. 
Robustness to such occlusions is critical for real-world applications.

Existing approaches~\cite{jiang2020coherent,kocabas2020vibe,zanfir2018monocular,zanfir2018deep} follow a multi-stage design that equips the single-person method with a 2D person detector to handle multi-person scenes.
Generally, they first detect regions with people and then extract the bounding-box-level features, which are used to regress each single 3D human mesh \cite{Guler_2019_CVPR,hmr,kanazawa2019learning,kocabas2020vibe,kolotouros2019spin,gcmr,pavlakos2019texturepose,sun2019dsd-satn,Xu_2019_ICCV,Zhu_2019_CVPR}. 
However, as shown in Fig.~\ref{fig:motivation}, this strategy is prone to fail in cases of multi-person occlusion and truncation. 
Specifically, as shown in Fig.~\ref{fig:motivation}(b), when two people overlap, it is hard for the multi-stage method to estimate diverse body meshes from similar image patches.
The ambiguity of the implicit bounding-box-level representation results in failure for such inseparable multi-person cases.

For multi-person 2D pose estimation, this problem has been tackled via a subtle and effective bottom-up framework. 
The paradigm first detects all body joints and then assigns them to different people by grouping body joints. 
This pixel-level body-joint representation enables their impressive performance in crowded scenes~\cite{openpose,cheng2020bottom,pishchulin2016deepcut}. 
However, it is non-trivial to extend this bottom-up one-stage process beyond joints~\cite{jiang2020coherent}.
Unlike 2D pose estimation, which predicts dozens of body joints, we need to regress a human body mesh with thousands of vertices, making it hard to follow the paradigm of body joint detection and grouping.

Instead, we introduce ROMP, a one-stage network for regressing multiple 3D people in a per-pixel prediction fashion.
It directly estimates multiple differentiable maps from the whole image, from which we can easily parse out the 3D meshes of all people.
Specifically, as shown in Fig.~\ref{fig:motivation}(c), ROMP predicts a Body Center heatmap and a Mesh Parameter map, representing the 2D position of the body center and the parameter vectors of the corresponding 3D body mesh, respectively.
Via a simple parameter sampling process, we extract 3D body mesh parameter vectors of all people from the Mesh Parameter map at the body center locations described by the heatmap. 
Then we put the sampled mesh parameter vectors into the SMPL body model~\cite{SMPL} to derive multi-person 3D meshes.

Following the guidance of body centers during training, the ambiguity of the regression target is greatly alleviated in crowded multi-person scenes. 
Additionally, in contrast to the local, bounding-box-level, features learned by traditional methods, end-to-end learning from the whole image forces the model to learn appropriate features from the holistic scene to predict bodies with occlusion.
This holistic approach captures the complexity of real-world scenes, enabling the generalization and robustness to complex multi-person cases.

Moreover, since the body centers of severely overlapping people may collide at the same 2D position, we further develop an advanced collision-aware representation (CAR). 
The key idea is to construct a repulsion field of body centers, where close body centers are analogous to positive charges that are pushed apart by mutual repulsion. In this way, the body centers of the overlapping people are more distinguishable. Especially in the case of severe overlap, most part of the human body is invisible. Mutual repulsion will push the center to the visible body area, meaning that the model tends to sample 3D mesh parameters estimated from the position centered on the visible body parts. It improves the robustness under heavy person-person occlusion. 

Compared with previous state-of-the-art methods for multi-person~\cite{jiang2020coherent,zanfir2018monocular,zanfir2018deep} and single-person~\cite{kocabas2020vibe,kolotouros2019spin} 3D mesh regression, ROMP 
achieves superior performance on challenging benchmarks, including 3DPW~\cite{3dpw} and CMU Panoptic~\cite{cmu_panoptic}. 
Experiments on person-person occlusion datasets (Crowdpose~\cite{crowdpose} and 3DPW-PC, a person-occluded subset of 3DPW~\cite{3dpw}) demonstrate the effectiveness of the proposed CAR under person-person occlusion.  
To further evaluate it in general cases, we test ROMP on images from the Internet and web camera videos. 
With the same backbone as the multi-stage counterparts, ROMP runs in real-time (over 30 FPS) on a 1070Ti GPU.

In summary, the contributions are:
(1) To the best of our knowledge, ROMP is the first one-stage method for monocular multi-person 3D mesh regression, along with an open-source real-time implementation. Its simple yet effective framework leads to superior accuracy and efficiency.
(2) The proposed explicit body-center-guided representation facilitates the pixel-level human mesh regression in an end-to-end manner.
(3) We develop a collision-aware representation to deal with cases of severe overlap.

\begin{figure*}[t]
	\centering
	\includegraphics[width=1.00\textwidth]{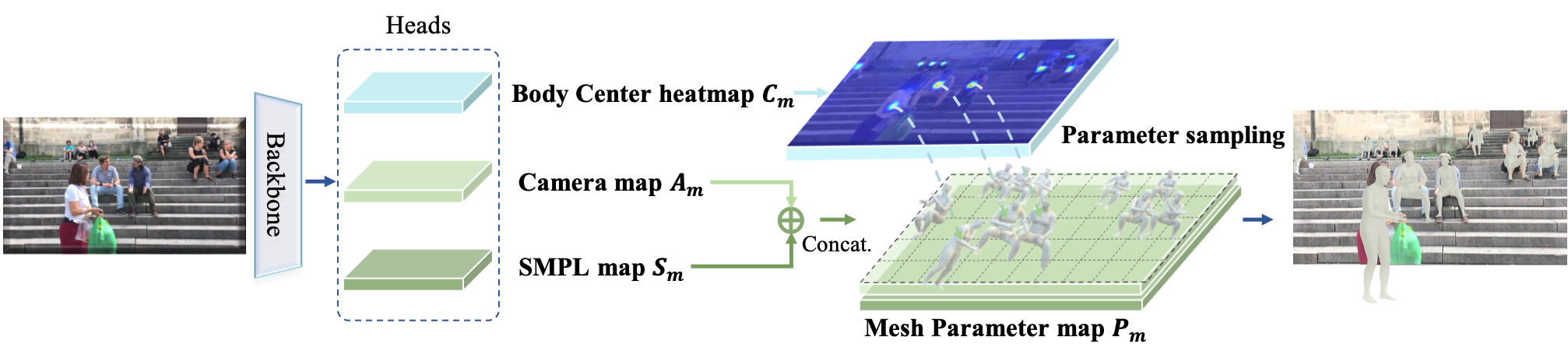}
	\caption{An overview of ROMP. Given an input image, ROMP predicts multiple maps: 1) the Body Center heatmap  predicts the probability of each position being a body center, 2) the Camera map and 3) SMPL map  contain the camera and SMPL parameters~\cite{SMPL} of the person at each center, respectively. As the concatenation of the Camera map and SMPL map, the Mesh Parameter map contains the information of the predicted 3D body mesh and its location. Via the designed parameter sampling process, we can obtain the final 3D mesh results by parsing the Body Center heatmap and sampling the Mesh Parameter map.}
	\label{fig:framework}
\end{figure*}

\section{Related Work}

\textbf{Single-person 3D mesh regression.} 
Parametric human body models, e.g., SMPL~\cite{SMPL}, have been widely adopted to encode the complex 3D human mesh into a low-dimensional parameter vector, which can be regressed from images~\cite{liu2021recent}.
Impressive performance has been achieved using various weak supervision signals, such as 2D pose~\cite{choi2020pose2mesh,hmr,kundu2020appearance}, semantic segmentation~\cite{Xu_2019_ICCV}, 
geometric priors and discriminative training~\cite{hmr}, 
motion dynamics~\cite{kanazawa2019learning}, temporal coherence~\cite{kocabas2020vibe,sun2019dsd-satn}, texture consistency~\cite{pavlakos2019texturepose}, optimization~\cite{keep} in the loop~\cite{kolotouros2019spin}, etc.
However, all these methods adopt a bounding-box-level representation, which is implicit and ambiguous for multi-person/occlusion cases.
To handle occluding objects, Zhang et al.~\cite{zhang2020object} use a 2D UV map to represent a 3D human mesh. 
Considering the object-occluded body parts as blank areas in the partial UV map, they in-paint the partial UV map to make up the occluded information.
However, in the case of person-person occlusion where one person's body parts are occluded by those of another, it is hard to generate the partial UV map.

\textbf{Multi-person 3D pose estimation}. 
Mainstream methods can be divided into two categories: the one-stage and the multi-stage.
Many multi-stage methods follow the top-down design of Faster R-CNN~\cite{ren2015fasterrcnn}, such as LCR-Net++~\cite{rogez2019lcr} and 3DMPPE~\cite{moon2019camera}. 
From anchor-based feature proposals, they estimate the target via regression.
Other works explore a one-stage solution that reasons about all people in a single forward pass. They estimate all the body joint positions and then group them into individuals.
Mehta et al.~\cite{mehta2018single} propose  occlusion-robust pose-maps and exploit the body part association to avoid the bounding box prediction.
PandaNet~\cite{PandaNet_Benzine_2020_CVPR} is an anchor-based one-stage model that relies on a huge number of pre-defined anchor predictions and the positive anchor selection. 
To handle person-person occlusion, Zhen et al.~\cite{zhen2020smap} extend part affinity fields \cite{openpose} to be depth-aware.
ROMP extends the end-to-end one-stage process beyond joints through a concise body-center-guided pixel-level representation and does not require the part association or an enormous number of anchor predictions.

\textbf{Multi-person 3D mesh regression.} 
There are only a few approaches for multi-person 3D mesh regression. 
Zanfir et al.~\cite{zanfir2018deep} estimate the 3D mesh of each person from its intermediate 3D pose estimation.
Zanfir et al.~\cite{zanfir2018monocular} further employ multiple scene constraints to optimize the multi-person 3D mesh results. 
Jiang et al.~\cite{jiang2020coherent} propose a network for Coherent Reconstruction of Multiple Humans (CRMH).
Built on Faster-RCNN~\cite{ren2015fasterrcnn}, they use the RoI-aligned feature of each person to predict the SMPL parameters. 
Additionally, they learn the relative positions between multiple people via an interpenetration and depth ordering loss.
All these methods follow a multi-stage design.
The complex multi-step process requires a repeated feature extraction, which is computationally expensive. 
Moreover, since they rely on detected bounding boxes, the ambiguity and the limited local view of the bounding-box-level features make it hard to effectively learn from person-person occlusion and truncation.
Instead, our proposed one-stage method learns an explicit pixel-level representation with a holistic view, which significantly improves both accuracy and efficiency in multi-person in-the-wild scenes. 
    
 \textbf{Pixel-level representations} have proven useful in anchor-free detection, such as CornerNet~\cite{law2018cornernet}, CenterNet~\cite{duan2019centernet,zhou2019objects}, and ExtremeNet~\cite{ExtremeNet}.
They directly estimate the corner or center point of the bounding box in a heatmap manner, which avoids the dense anchor-based proposal.
Inspired by these, we develop a pixel-level fine-grained representation for multi-person 3D meshes.
Different from the bounding box center used in~\cite{zhou2019objects}, our body center is determined by the body joints, as introduced in Sec.~\ref{sec:CAR}. 
A recent work, BMP~\cite{zhang2021bmp}, uses a multi-scale grid-level representation for multi-person 3D mesh recovery, which locates a target person at the center of the grid cell.\footnote{The arXiv version of ROMP, called CenterHMR~\cite{centerhmr}, predates BMP.}
In contrast to these methods, ROMP adopts a concise body-center-based representation and further evolves it to a collision-aware version to deal with the inherent center collision problem.

\textbf{Disambiguation} is a key goal of ROMP, enabling it to  deal with the crowded multi-person scenes. Related techniques have been studied in many other fields. For instance segmentation, Adaptis~\cite{sofiiuk2019adaptis} separately learns the segmentation mask of each instance selected by the guide point. 
To alleviate the ambiguity between embeddings of similar samples, associate embedding~\cite{newell2017associative}, triplet loss~\cite{tripletloss}, and pose-guided association~\cite{bao2020pose} are developed for pose estimation, face recognition, and tracking respectively. 
In this paper, a robust and distinguishable representation is developed to help the model explicitly learn from the crowded scenes.

\section{Our Approach}

\vspace{-1mm}
\subsection{Overview}

The overall framework is illustrated in Fig.~\ref{fig:framework}.
It adopts a simple multi-head design with a backbone and three head networks.
Given a single RGB image as input, it outputs a Body Center heatmap, Camera map, and SMPL map, describing the detailed information of the estimated 3D human meshes. 
In the Body Center heatmap, we predict the probability of each position being a human body center. 
At each position of the Camera/SMPL map, we predict the camera/SMPL parameters of the person that takes the position as the center. 
For simplicity, we combine the Camera map and SMPL map into the Mesh Parameter map. 
During inference, we sample the 3D body mesh parameter results from the Mesh Parameter map at the 2D body center locations parsed from the Body Center heatmap. 
Finally, we put the sampled parameters into the SMPL model to generate the 3D body meshes. 

\subsection{Basic Representations~\label{sec:representations}}

In this section, we introduce the detailed representation of each map. Each output map is of size $n \times H \times W$, where $n$ is the number of channels. Here, we set $H=W=64$. 

\textbf{Body Center heatmap:} $\boldsymbol{C_m} \in \mathbb{R}^{1 \times H \times W}$ is a heatmap representing the 2D human body center in the image. 
Each body center is represented as a Gaussian distribution in the Body Center heatmap. 
For better representation learning, the Body Center heatmap also integrates the scale information of the body in the 2D image.
Specifically, we calculate the Gaussian kernel size $k$ of each person center in terms of its 2D body scale in the image. 
Given the diagonal length $d_{bb}$ of the person bounding box and the width $W$ of the Body Center heatmap, $k$ is derived as
\begin{equation}
\setlength{\abovedisplayskip}{0pt}
\setlength{\belowdisplayskip}{1pt}
\begin{aligned}
k = k_l+(\frac{ d_{bb}}{\sqrt{2}W})^2 k_r ,
\end{aligned}
\label{eq:center kernel size}
\end{equation}
where $k_l$ is the minimum kernel size and $k_r$ is the variation range of $k$. We set $k_l=2$ and  $k_r=5$ by default.

\textbf{Mesh Parameter map:} $\boldsymbol{P_m} \in \mathbb{R}^{145 \times H \times W}$ consists of two parts, the Camera map and SMPL map.
Assuming that each location of these maps is the center of a human body, we estimate its corresponding 3D body mesh parameters.
Following~\cite{hmr,sun2019dsd-satn}, we employ a weak-perspective camera model to project $K$ 3D body joints $\boldsymbol{J}=(x_k,y_k,z_k), k=1\cdots K$ of the estimated 3D mesh back to the 2D joints $\boldsymbol{\widehat J}=(\widehat x_k,\widehat y_k)$ on the image plane. 
This facilitates training the model with in-the-wild 2D pose datasets, which helps with robustness and generalization. 
 
\textbf{Camera map:} $\boldsymbol{A_m} \in \mathbb{R}^{3 \times H \times W}$ contains the 3-dim camera parameters $(s, t_x, t_y)$ that describe the 2D scale $s$ and translation $\boldsymbol{t}=(t_x, t_y)$ of the person in the image. 
The scale $s$ reflects the body size and the depth to some extent.
$t_x$ and $t_y$, ranging in $(-1,1)$, reflect the normalized translation of the human body relative to the image center on the $x$ and $y$ axis, respectively.
The 2D projection $ \boldsymbol{\widehat J}$ of  3D body joints $\boldsymbol{J}$ can be derived as  $ \widehat x_k = s x_k+t_x, \widehat y_k = s y_k+t_y$.
The translation parameters allow more accurate position estimates than the Body Center heatmap. 

\textbf{SMPL map:} $\boldsymbol{S_m} \in \mathbb{R}^{142 \times H \times W}$  contains the 142-dim SMPL parameters, which describe the 3D pose and shape of the body mesh. 
SMPL establishes an efficient mapping from the pose  $ \boldsymbol{\theta}$  and shape $ \boldsymbol{\beta}$ parameters to the human 3D body mesh $\boldsymbol{M}\in\mathbb{R}^{6890 \times 3}$.
The shape parameter $ \boldsymbol{\beta} \in \mathbb{R}^{10}$ is the top-10 PCA coefficients of the SMPL statistical shape space. 
The pose parameters $ \boldsymbol{\theta} \in \mathbb{R}^{6 \times 22}$ contain the 3D rotation of the 22 body joints in a 6D representation~\cite{Zhou_2019_CVPR}.  
Instead of using the full 24 joints of the original SMPL model, we drop the last two hand joints. 
The 3D rotation of the first joint denotes the body 3D orientation in the camera coordinate system, 
while the remainder are the relative 3D orientations of each body part with respect to its parent in a kinematic chain. 
3D joints $\boldsymbol{J}$ are derived via $\boldsymbol{P}\boldsymbol{M}$, where $\boldsymbol{P} \in \mathbb{R}^{K \times 6890}$ is a sparse weight matrix that describes the linear mapping from 6890 vertices of the body mesh $\boldsymbol{M}$ to the $K$ body joints.

\subsection{CAR: Collision-Aware Representation~\label{sec:CAR}}

\begin{figure}[t]
	\centering
	\includegraphics[width=0.98\columnwidth]{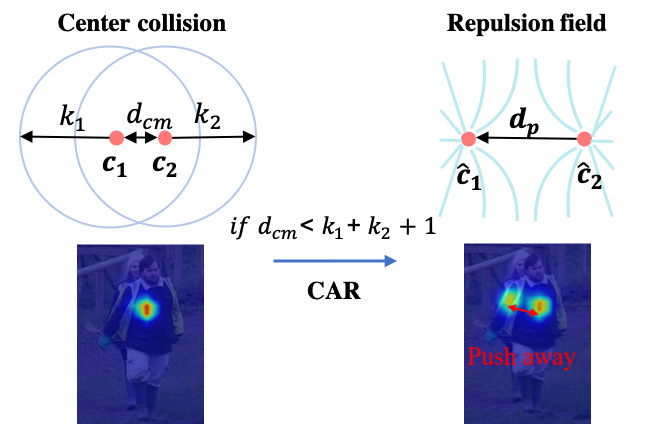} 
	\caption{{\bf Collision-Aware Representation.} The body centers of overlapping people are treated as positive charges, which are pushed apart if they are too close in the repulsion field. }
	\label{fig:CAR}
\end{figure}

The entire framework is based on a concise body-center-guided representation. 
It is crucial to define an explicit and robust body center so that the model can easily estimate the center location in various cases. 
Here we introduce the basic definition of the body center for the general case and its advanced version for severe occlusion.

\textbf{Basic definition of the body center.} Existing center-based methods~\cite{duan2019centernet,zhou2019objects} define the center of the bounding box as the target center. 
This works well for general objects (e.g., a ball or bottle) that lack semantically meaningful keypoints.
However, a bounding box center is not a meaningful point on the human body and often falls outside the body area.
For stable parameter sampling, we need an explicit body center.
Therefore, we calculate each body center from the ground truth 2D pose. 
Considering that any body joint may be occluded in general cases, we define the body center as the center of visible torso joints (neck, left/right shoulders, pelvis, and left/right hips). 
When all torso joints are invisible, the center is simply determined by the average of the visible joints.
In this way, the model is encouraged to predict the body location from the visible parts. 

However, in cases of severely overlapping people, the body center of the people might be very close or even at the same location on $\boldsymbol{C_m}$.
This center collision problem makes the center ambiguous and hard to identify in crowded cases.
To tackle this, we develop a more robust representation to deal with person-person occlusion.
To alleviate the ambiguity, the center points of overlapping people should be kept at a minimum distance to ensure that they can be well distinguished. 
Additionally, to avoid sampling multiple parameters for the same person, the network should assign a unique and explicit center for each person.

Based on these principles, we develop a novel \textbf{Collision-Aware Representation (CAR)}. 
To ensure that the body centers are far enough from each other, we construct a repulsion field. 
In this field, each body center is treated as a positive charge, whose radius of repulsion is equal to its Gaussian kernel size derived by Eq.~(\ref{eq:center kernel size}).
In this way, the closer the body centers are, the greater the mutual repulsion and the further they will be pushed apart. 
Fig. \ref{fig:CAR} illustrates the principle of CAR.  
Suppose that $\boldsymbol{c_1} \in\mathbb{R}^{2}, \boldsymbol{c_2} \in\mathbb{R}^{2}$ are the  body centers of two overlapping people. 
If  their Euclidean distance $d_{cm}$ and Gaussian kernel sizes $k_1, k_2$ satisfy  $d_{cm}<k_1+k_2+1$, the repulsion is triggered to push the close centers apart  via
\begin{equation}
\setlength{\abovedisplayskip}{2pt}
\setlength{\belowdisplayskip}{2pt}
\begin{aligned}
&\boldsymbol{\hat{c}_1} = \boldsymbol{c_1} + \gamma \boldsymbol{d_{p}}, \boldsymbol{\hat{c}_2} = \boldsymbol{c_2} - \gamma \boldsymbol{d_{p}},\\
&\boldsymbol{d_{p}} = \frac{k_1+k_2+1 - d_{cm}}{d_{cm}}  (\boldsymbol{c_1}-\boldsymbol{c_2}),
\end{aligned}
\label{eq:centerpush}
\end{equation}
where $\boldsymbol{d_{p}}\in \mathbb{R}^{2}$ is the repulsion vector from $\boldsymbol{c_2}$ to $\boldsymbol{c_1}$ and $\gamma$ is an intensity coefficient to adjust the strength. When there are  multiple overlapped people, we take Eq.~(\ref{eq:centerpush}) to generate the mutual repulsion vectors $\boldsymbol{d^i_{p}}$ for the $i$-th pair of centers. For the center that is affected by $N$  repulsive forces, we calculate the composition of these forces as the numerical summation $\sum_{i=1}^{N} \boldsymbol{d^i_p}$.

During training, we use CAR to push apart close body centers and use them to supervise the Body Center heatmap. 
In this way, the model is encouraged to estimate the  centers that maintain a distinguishable distance.
For the Body Center heatmap, it helps the model to effectively locate the occluded person.
For the Mesh Parameter map, sampling the parameters from these shifted locations enables the model to extract diverse and individual features for each person.
The model trained with CAR is more suitable for crowded scenes with significant person-person occlusion such as train stations, canteens, etc.

\subsection{Parameter Sampling~\label{sec:parameter sampling}}

To parse the 3D body meshes from the estimated maps, we need to first parse the 2D body center coordinates $\boldsymbol{c} \in \mathbb{R}^{K \times 2}$ from $\boldsymbol{C_m}$, where $K$ is the number of the detected people, and then use them to sample $\boldsymbol{P_m}$ for the SMPL parameters. 
In this section, we introduce the process of the center parsing, matching and sampling. 

$\boldsymbol{C_m}$ is a probability map whose local maxima are regarded as the body centers. 
The local maxima are derived via $Mp(\boldsymbol{C_m}) \land \boldsymbol{C_m}$ where $Mp$ is the max pooling operation and $\land$ is the logical conjunction operation. 
Let $\boldsymbol{c}$ be the 2D coordinates of a local maximum with confidence score larger than a threshold $t_c$. 
We rank the confidence score at each $\boldsymbol{c}$ and take the top $N$ as the final centers.
During inference, we directly sample the parameters from $\boldsymbol{P_m}$ at $\boldsymbol{c}$. 
During training, the estimated $\boldsymbol{c}$ are matched with the nearest ground truth body center according to the $L_2$ distance. 

Additionally, we approximate the depth order between multiple people by using the center confidence from $\boldsymbol{C_m}$ and the 2D body scale $s$ of the camera parameters from $\boldsymbol{A_m}$.
For people of different $s$, we regard the one with the larger $s$ as located in the front.
For people of similar $s$, the person with a higher center confidence is considered to be in the front. Please refer to the SuppMat for the details.

\subsection{Loss Functions}

To supervise ROMP, we develop individual loss functions for different maps.
In total, ROMP is supervised by the weighted sum of the body center loss $L_{c}$ and mesh parameter loss $L_{p}$.

\textbf{Body Center loss.} $L_{c}$ encourages a high confidence value at the body center $\boldsymbol{c}$ of the Body Center heatmap $\boldsymbol{C_m}$ and low confidence elsewhere.
To deal with the imbalance between the center location and the non-center locations in $\boldsymbol{C_m}$, we train the Body Center heatmap based on the focal loss~\cite{lin2017focal}. 
Given the predicted Body Center heatmap $\boldsymbol{C_{m}^p}$ and the ground truth $\boldsymbol{C_{m}^{gt}}$,  $L_{c}$ is defined as 
\begin{equation}
\setlength{\abovedisplayskip}{5pt}
\setlength{\belowdisplayskip}{6pt}
\begin{aligned}
&L_{c} = - \frac{L_{pos}+L_{neg}}{\sum \boldsymbol{I_{pos}}} w_{c},\\
&L_{neg} = log(1-\boldsymbol{C_{m}^p})(\boldsymbol{C_{m}^p})^2 (1-\boldsymbol{C_{m}^{gt}})^4 (1-\boldsymbol{I_{pos}}),\\
&L_{pos} = log(\boldsymbol{C_{m}^p})(1-\boldsymbol{C_{m}^p})^2 \boldsymbol{I_{pos}},  \boldsymbol{I_{pos}}=\boldsymbol{C_{m}^{gt}}\geq1,
\end{aligned}
\label{eq:focal loss}
\end{equation}
where $\boldsymbol{I_{pos}}$  is a binary matrix with a positive value at the body center location, and $w_{c}$ is the loss weight.

\textbf{Mesh Parameter loss.} 

As we introduced in Sec.~\ref{sec:parameter sampling}, the parameter sampling process matches each ground truth body with a predicted parameter result for supervision.
The mesh parameter loss is derived as 
\begin{equation}
\setlength{\abovedisplayskip}{4pt}
\setlength{\belowdisplayskip}{5pt}
\begin{aligned}
L_p = w_{pose}L_{pose}+w_{shape}L_{shape}+w_{j3d}L_{j3d}\\+w_{paj3d}L_{paj3d}+w_{pj2d}L_{pj2d}+w_{prior}L_{prior}.
\end{aligned}
\label{eq:parameter loss}
\end{equation}
$L_{pose}$ is the $L_2$ loss of the pose parameters in the $3 \times 3$ rotation matrix format. 
$L_{shape}$ is the $L_2$ loss of the shape parameters. 
$L_{j3d}$ is the $L_2$ loss of the 3D joints $\boldsymbol{J}$ regressed from the body mesh $\boldsymbol{M}$.
$L_{paj3d}$ is the $L_2$ loss of the 3D joints $\boldsymbol{J}$ after Procrustes alignment.
$L_{pj2d}$ is the $L_2$ loss of the projected 2D joints $\boldsymbol{\widehat J}$. 
$L_{prior}$ is the Mixture Gaussian prior loss of the SMPL parameters adopted in~\cite{keep,SMPL} for supervising the plausibility of 3D joint rotation and body shape. Lastly, $w_{(.)}$ denotes the corresponding loss weights.

\begin{table*}[t]
\setlength\tabcolsep{2pt}
  \centering
  {
  \rowcolors{2}{gray!10}{white}
    \begin{tabular}{l|cccccc}
    \hline
    \textbf{Method} & \textbf{MPJPE$\downarrow$} &  \textbf{PMPJPE$\downarrow$} &\textbf{PCK$\uparrow$} &	\textbf{AUC$\uparrow$} &\textbf{MPJAE$\downarrow$} & \textbf{PMPJAE$\downarrow$} \\
    \hline
        OpenPose + SPIN~\cite{kolotouros2019spin}&  95.8 &  66.4 &33.3 & 55.0  & 23.9 & 24.4\\
        YOLO + VIBE~\cite{kocabas2020vibe}$^\star$& 94.7 & 66.1 & 33.9 & 56.6 &25.2 &20.46 \\
        CRMH~\cite{jiang2020coherent}& 105.9 &71.8 & 28.5 & 51.4 & 26.4 & 22.0\\
        BMP~\cite{zhang2021bmp}$^\star$ & 104.1 & 63.8 & 32.1 & 54.5 & - & - \\
        \hline
        ROMP (ResNet-50) & 87.0 & 62.0 & 34.4 & 57.6 & 21.9 & 20.1\\
        ROMP (ResNet-50)$^\star$ & \textbf{80.1} &  \textbf{56.8} &36.4 & \textbf{60.1} & 20.8 &19.1\\
        ROMP (HRNet-32)$^\star$ &82.7 & 60.5 &  \textbf{36.5} & 59.7 & \textbf{20.5} & \textbf{18.9}\\
    \hline
    \end{tabular} }
    \caption{{Comparisons to the state-of-the-art methods  on 3DPW following \textit{Protocol 1} (without using any ground truth during inference). $^\star$ means using extra datasets for training.}}  \label{tab:3DPW}%
\end{table*}

\begin{table}
 \setlength{\tabcolsep}{2pt}{
	\begin{center}
        \rowcolors{2}{gray!10}{white}
			\begin{tabular}{l|ccc}
				\hline
			    \textbf{Method}  &  \textbf{MPJPE$\downarrow$} & \textbf{PMPJPE$\downarrow$}  & \textbf{PVE$\downarrow$}\\
				\hline 
                 HMR~\cite{hmr} & 130.0 & 76.7 & - \\
                 Kanazawa et al.~\cite{kanazawa2019learning}  & 116.5 & 72.6 & 139.3 \\
                 Arnab et al.~\cite{arnab2019exploiting}  & - & 72.2 & - \\
                 GCMR~\cite{gcmr} & - & 70.2 & - \\
                 DSD-SATN~\cite{sun2019dsd-satn}  & - & 69.5 & - \\
                 SPIN~\cite{kolotouros2019spin}  & 96.9 & 59.2 & 116.4 \\
                 ROMP (ResNet-50) & \textbf{91.3} & \textbf{54.9} & \textbf{108.3}\\
                 \hline
                  I2L-MeshNet~\cite{moon2020i2l}$^\star$ & 93.2 & 58.6 & - \\
                 EFT~\cite{joo2020eft}$^\star$  & - & 54.2 & - \\
                 VIBE~\cite{kocabas2020vibe}$^\star$ & 93.5 & 56.5 & 113.4\\
                 ROMP (ResNet-50)$^\star$& 89.3 & 53.5 & 105.6\\
                 ROMP (HRNet-32)$^\star$ &\textbf{85.5} & \textbf{53.3} & \textbf{103.1}\\
				\hline
		\end{tabular}
	\end{center}}
    \caption{{Comparisons to the state-of-the-art methods on 3DPW following VIBE~\cite{kocabas2020vibe}, using \textit{Protocol 2} (on the test set only). $^\star$ means using extra datasets (compared with SPIN) for training.}}\label{tab:3DPW_VIBE2}
\end{table}

\section{Experiments}
\vspace{-1mm}
\subsection{Implementation Details}

\textbf{Network Architecture.} 
For a fair comparison with other approaches, we use ResNet-50~\cite{resnet} as the default backbone.
Since our method is not limited to a specific backbone, we also test HRNet-32~\cite{cheng2020bottom} in the experiments.
Through the backbone, a feature vector $\boldsymbol{f_b} \in \mathbb{R}^{32 \times H_b \times W_b}$ is extracted from a single RGB image. 
Also, we adopt the CoordConv~\cite{liu2018coordconv} to enhance the spatial information.
Therefore, the backbone feature $\boldsymbol{f} \in \mathbb{R}^{34 \times H_b \times W_b}$ is the combination of a coordinate index map $\boldsymbol{ci} \in \mathbb{R}^{2 \times H_b \times W_b}$ and $\boldsymbol{f_b}$.
Next, from $\boldsymbol{f}$, three head networks are developed to estimate the Body Center, Camera, and SMPL maps. 
More details of the architecture are in the SuppMat.

\textbf{Setting Details.} 
The input images are resized to $512 \times 512$, keeping the same aspect ratio and padding with zeros. 
The size of the backbone feature is $H_b=W_b=128$. 
The maximum number of detections, $N=64$, which is set manually. 
The loss weights are set to $w_{c}=160,w_{j3d}=360,w_{paj3d}=400,w_{pj2d}=420,w_{pose}=80,w_{shape}=1$, and $w_{prior}=1.6$ to ensure that the weighted loss items are of the same magnitude.
The threshold $t_c$ of the Body Center heatmap is 0.2. The repulsion coefficient $\gamma$ of CAR is 0.2. 

\textbf{Training Datasets.}
For a fair comparison with previous methods~\cite{jiang2020coherent,hmr,kolotouros2019spin,sun2019dsd-satn}, the basic training datasets we used in the experiments include two 3D pose datasets (Human3.6M~\cite{h36m} and MPI-INF-3DHP~\cite{mono-3dhp2017}), one pseudo-label 3D dataset (UP~\cite{unite}) and four in-the-wild 2D pose datasets (MS COCO~\cite{coco}, MPII~\cite{mpii}, LSP~\cite{lsp,lsp_extended} and AICH~\cite{aich}). 
We also use the pseudo 3D annotations from \cite{kolotouros2019spin}. 
To further explore the upper limit of performance, we also use additional training datasets, including two 3D pose datasets (MuCo-3DHP~\cite{mono-3dhp2017} and OH~\cite{zhang2020object}), the pseudo 3D labels of 2D pose datasets provided by~\cite{joo2020eft}, and two 2D pose datasets (PoseTrack~\cite{PoseTrack} and Crowdpose~\cite{crowdpose}), to train an advanced model.

\textbf{Evaluation Benchmarks.} 
3DPW~\cite{3dpw} is employed as the main benchmark for evaluating 3D mesh/joint error since it contains in-the-wild multi-person videos with abundant 2D/3D annotations.
Specially, we divide 3DPW into 3 subsets, including 3DPW-PC for person-person occlusion, 3DPW-OC for object occlusion, and 3DPW-NC for the non-occluded/truncated cases, to evaluate the performance in different scenarios.
Additionally, we also evaluate on a indoor multi-person 3D pose benchmark, CMU Panoptic~\cite{cmu_panoptic}. 
Furthermore, we evaluate the stability under occlusion on  Crowdpose~\cite{crowdpose}, a crowded-person in-the-wild 2D pose benchmark.

\textbf{Evaluation Metrics.}
We adopt per-vertex error (PVE) to evaluate the 3D surface error.
To evaluate the 3D pose accuracy, we employ mean per joint position error (MPJPE), Procrustes-aligned MPJPE (PMPJPE), percentage of correct keypoints (PCK), and area under the PCK-threshold curve (AUC).
We also adopt mean per joint angle error (MPJAE), and Procrustes-aligned MPJAE (PA-MPJAE) to evaluate the 3D joint rotation accuracy. 
Also, to evaluate the pose accuracy in crowded scenes, we calculate the average precision ($ \rm{AP}^{0.5}$) between the 2D-projection $\boldsymbol{\widehat J}$ and the ground truth 2D poses on Crowdpose.

\begin{table}
 \setlength{\tabcolsep}{2pt}{
	\begin{center}
        \rowcolors{2}{gray!10}{white}
			\begin{tabular}{l|ccc}
				\hline
			    \textbf{Method} &  \textbf{MPJPE$\downarrow$} & \textbf{PMPJPE$\downarrow$}  &\textbf{PVE$\downarrow$}\\
				\hline 
                 EFT~\cite{joo2020eft} &   - & 52.2 & - \\
                 VIBE~\cite{kocabas2020vibe}$^\star$ & 82.9 & 51.9 & 99.1\\
                 \hline 
                 ROMP (ResNet-50) &  84.2 & 51.9 & 100.4 \\
                 ROMP(ResNet-50)$^\star$ & 79.7 & 49.7 & 94.7 \\
                 ROMP (HRNet-32)   & 78.8 & 48.3 & 94.3 \\
                 ROMP (HRNet-32)$^\star$ & \textbf{76.7} & \textbf{47.3} & \textbf{93.4}\\
				\hline
		\end{tabular}
	\end{center}}
    \caption{ {Comparisons to the state-of-the-art methods on 3DPW following \textit{Protocol 3} (fine-tuned on the training set). $^\star$ means using extra datasets (compared with EFT) for training.}}\label{tab:3DPW_VIBE3}
\end{table}

\begin{table}[t]
\setlength\tabcolsep{0.5mm}
  \centering
  {
    \rowcolors{2}{gray!10}{white}
    \begin{tabular}{l|cccc|c}
    \hline
    \textbf{Method} &  \textbf{Haggling} & \textbf{Mafia} &\textbf{Ultim.} & \textbf{Pizza} & \textbf{Mean}\\
    \hline
        Zanfir et. al.~\cite{zanfir2018deep} & 141.4 & 152.3 & 145.0 & 162.5 & 150.3 \\
        MSC~\cite{zanfir2018monocular} & 140.0 & 165.9 & 150.7 & 156.0 & 153.4 \\
        CRMH~\cite{jiang2020coherent} & 129.6 & 133.5 & 153.0 & 156.7 & 143.2 \\
	    ROMP (ResNet-50) & \textbf{111.8} & \textbf{129.0} & \textbf{148.5} & \textbf{149.1} & \textbf{134.6} \\
        \hline
        BMP~\cite{zhang2021bmp}$^\star$ & 120.4 & 132.7 & 140.9 & 147.5 & 135.4\\
        ROMP (ResNet-50)$^\star$ & \textbf{107.8} & 125.3 & \textbf{135.4} & 141.8 & \textbf{127.6} \\
        ROMP (HRNet-32)$^\star$ & 110.8 & \textbf{122.8} & 141.6 & \textbf{137.6} & 128.2 \\
    \hline
    \end{tabular} }
    \caption{ {Comparisons to the state-of-the-art methods on CMU Panoptic~\cite{cmu_panoptic} benchmark. The evaluation metric is MPJPE after centering the root joint. All methods are directly evaluated without any fine-tuning.  $^\star$ means using extra datasets for training.}}  \label{tab:CMU Panoptic}
\end{table}

\begin{table}
	\begin{center}
        \setlength{\tabcolsep}{0.8mm}{
        {
          \rowcolors{2}{gray!10}{white}
			\begin{tabular}{l|ccc}
				\hline
				\textbf{Method} & \textbf{3DPW-PC} &  \textbf{3DPW-NC} & \textbf{3DPW-OC} \\
				\hline 
                CRMH~\cite{jiang2020coherent} & 103.5  & 68.7 & 78.9\\
                VIBE~\cite{kocabas2020vibe}  &103.9  & 57.3 &\textbf{ 65.9}\\
                \hline 
                ROMP w/o CAR & 79.7 & 56.7 &  67.0\\
                - w/ CAR ($\gamma=0.1$) & 77.6 & \textbf{55.6} & 66.6 \\
                - w/ CAR ($\gamma=0.2$) & \textbf{75.8}  & 57.1 & 67.1\\
                - w/ CAR ($\gamma=0.3$) & 77.0 & 56.4 & 66.5 \\
				\hline
		\end{tabular}}}
	\end{center}
	\caption{ {Comparisons to the state-of-the-art methods on the person-occluded (3DPW-PC),  object-occluded (3DPW-OC) and  non-occluded/truncated (3DPW-NC) subsets of 3DPW. We also ablate CAR and vary the  repulsion coeff.~$\gamma$.  The evaluation metric is PMPJPE. }}\label{tab:3DPW-SPLIT}
\end{table}

\subsection{Comparisons to the State-of-the-Art~\label{sec:comparisons}}

\textbf{3DPW.} 
We adopt three evaluation protocols, which reveal different properties.
To validate the performance in actual scenes, we follow \textit{Protocol 1} from the 3DPW Challenge to evaluate on the entire 3DPW dataset without using any ground truth, especially the bounding box. 
With the whole image as input, we equip each single-person method~\cite{kocabas2020vibe,kolotouros2019spin} with a human detector (OpenPose~\cite{openpose} or YOLO~\cite{redmon2018yolov3}).
For a fair comparison, ROMP uses the same backbone (ResNet-50) and training datasets as the competing method~\cite{kolotouros2019spin}.
We obtain the results of  OpenPose + SPIN from~\cite{imry2020challengespin}. 
The results of YOLO + VIBE are obtained using the officially released code, which already contains the YOLO part for human detection.
The results of BMP, which adopts a multi-scale grid-level representation, are obtained from~\cite{zhang2021bmp}.
In Tab.~\ref{tab:3DPW}, ROMP significantly outperforms all these methods, particularly in MPJPE, PMPJPE, and MPJAE.
These results validate that learning a robust pixel-level representation with a holistic view is helpful for improving the robustness and generalization in actual scenes.
Training with extra datasets ($*$) shows that the accuracy of ROMP can be further improved.

As a sanity check, we also compare ROMP with the single-person approaches in the evaluation protocols that allow them to use the cropped single-person image as input, while ROMP still takes the whole image as input.
Following VIBE~\cite{kocabas2020vibe}, \textit{Protocol 2} uses the 3DPW test set for evaluation without fine-tuning on the training set, while \textit{Protocol 3} fine-tunes the model on the 3DPW training set and uses the test set for evaluation.
In Tab.~\ref{tab:3DPW_VIBE2}, ROMP outperforms these multi-stage approaches on \textit{Protocol 2}, further demonstrating the advantage of our one-stage design.
In Tab.~\ref{tab:3DPW_VIBE3}, ROMP achieves comparable results with the state-of-the-art methods. 
If we use HRNet-32 as the backbone, the accuracy improves significantly after fine-tuning. 

\begin{table}
	\begin{center}
		
        \setlength{\tabcolsep}{0.6mm}{
			\begin{tabular}{l|c>{\columncolor{gray!10}}ccc}
                 \hline
				\textbf{Split}  & CRMH~\cite{jiang2020coherent}& ROMP &  ROMP+CAR \\
				\hline 
               Test & 33.9 & 54.1 & \textbf{59.7}\\
               Validation  & 32.9& 55.6 & \textbf{58.6}\\
				\hline
		\end{tabular}}
	\end{center}
   \caption{{Comparisons to the state-of-the-art methods on the Crowdpose~\cite{crowdpose} benchmark.  The evaluation metric is $ \rm{AP}^{0.5}$.}}\label{tab:Crowdpose}
\end{table}

\begin{table}
	\begin{center}
        \setlength{\tabcolsep}{0.6mm}{
			\begin{tabular}{c|c>{\columncolor{gray!10}}cc>{\columncolor{gray!10}}c}
				\hline
				\textbf{Method} & VIBE~\cite{kocabas2020vibe} & CRMH~\cite{jiang2020coherent}&ROMP& ROMP \\
                \hline
                FPS $\uparrow$ & 10.9 & 14.1 & 20.8 & \textbf{30.9}\\
                \hline
                Backbone & ResNet-50 & ResNet-50 & HRNet-32 & ResNet-50\\
				\hline
		\end{tabular}}
	\end{center}
	\caption{{Run-time comparisons on a 1070Ti GPU. }}\label{tab:runtime comparison}
\end{table}

\begin{figure}[t]
	\centering
	\includegraphics[width=0.98\columnwidth]{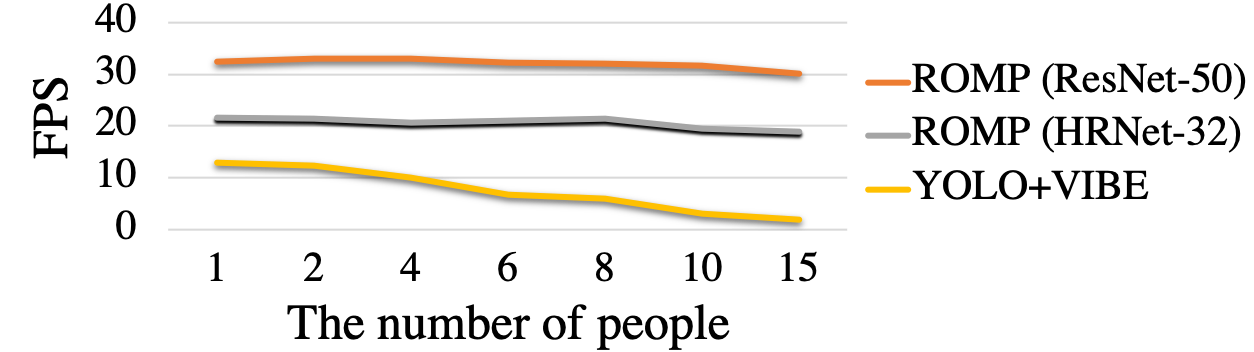}
	\caption{ The FPS variations of ROMP and YOLO+VIBE~\cite{kocabas2020vibe} when processing images with different number of people.}
	\label{fig:fps}
\end{figure}

\textbf{CMU Panoptic.} 
Following the evaluation protocol of CRMH~\cite{jiang2020coherent}, we evaluate ROMP on the multi-person benchmark, CMU Panoptic, without any fine-tuning.
For a fair comparison, we use the same backbone and similar training set as CRMH.
As shown in Tab.~\ref{tab:CMU Panoptic} , ROMP outperforms the existing multi-stage  methods~\cite{jiang2020coherent,zanfir2018monocular,zanfir2018deep} in all activities by a large margin.
These results further demonstrate that learning pixel-level representation with a holistic view improves the performance on multi-person scenes.

\begin{figure}[t]
	\centering
	\includegraphics[width=0.96\columnwidth]{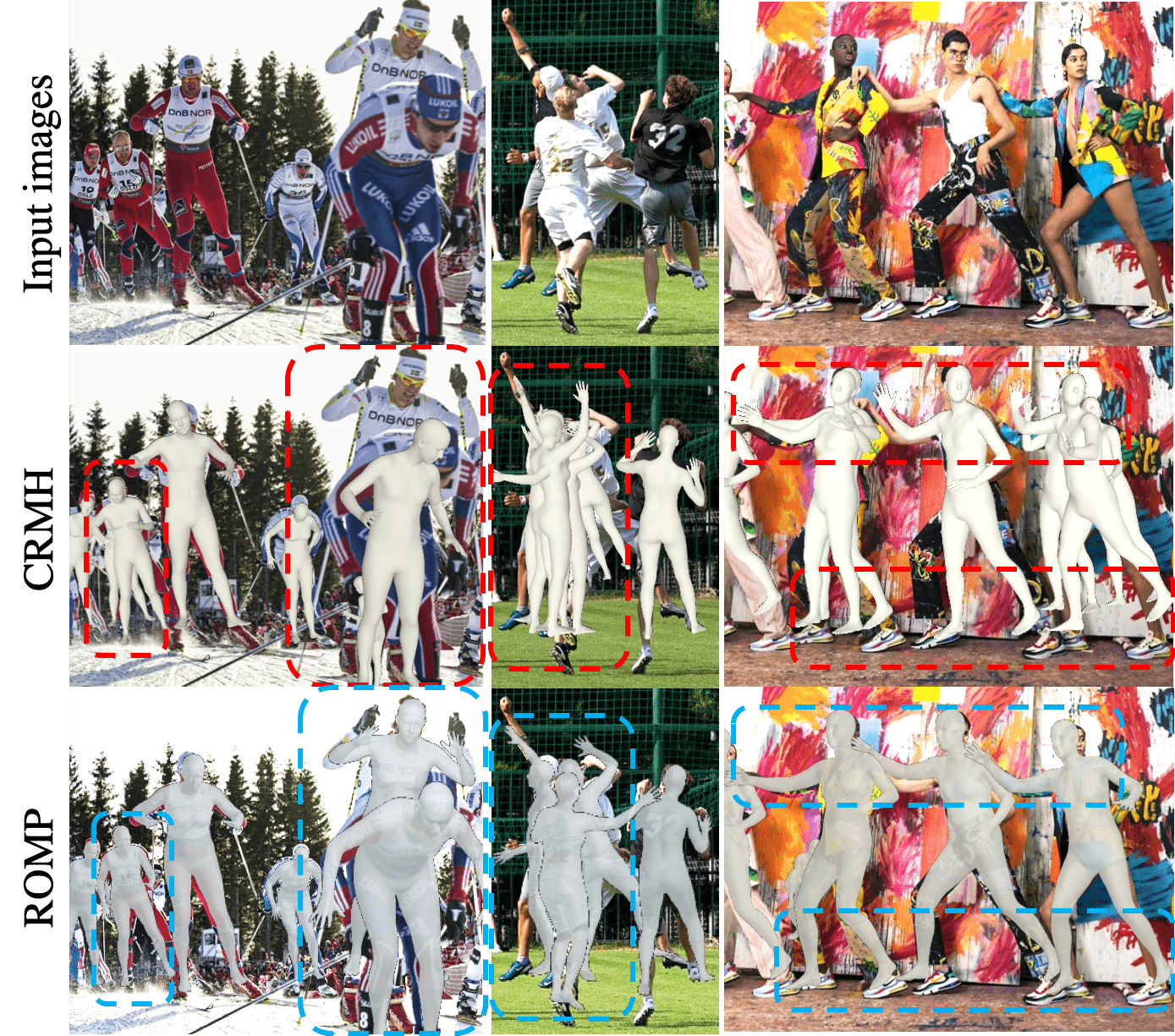}
	\caption{Qualitative comparisons to CRMH~\cite{jiang2020coherent} on the Crowdpose and the internet images.} 
	\label{fig:demo_comparison}
\end{figure}

\textbf{Occlusion benchmarks.}
To validate the stability under occlusion, we evaluate ROMP on multiple occlusion benchmarks.
Firstly, on the person-occluded \textbf{3DPW-PC} and \textbf{Crowdpose}~\cite{crowdpose}, results in Tab.~\ref{tab:3DPW-SPLIT} and \ref{tab:Crowdpose} show that ROMP significantly outperforms previous state-of-the-art methods~\cite{jiang2020coherent,kocabas2020vibe}.
Additionally, in Fig.~\ref{fig:demo_comparison}, some qualitative comparisons to 
CRMH also demonstrate ROMP's robustness to person-person occlusion.
These results suggest that the pixel-level representation is important for improving the performance under person-person occlusion.
Finally, on the object-occluded \textbf\textbf{3DPW-OC}, ROMP also achieves promising performance.
These results demonstrate that the fine-grained pixel-level representation is beneficial for dealing with various occlusion cases.

\textbf{Runtime comparisons.} 
We compare ROMP with the state-of-the-art methods in processing videos captured by a web camera.
All runtime comparisons are performed on a desktop with a GTX 1070Ti GPU, Intel i7-8700K CPU, and 8 GB RAM.
As shown in Tab.~\ref{tab:runtime comparison}, ROMP achieves real-time performance, significantly faster than the competing methods.
Additionally, as shown in Fig.~\ref{fig:fps}, compared with the multi-stage methods~\cite{jiang2020coherent,kocabas2020vibe}, ROMP's processing time is roughly constant regardless of the number of people. 

\vspace{-1mm}
\subsection{Ablation Study of the CAR}
\begin{figure}[t]
	\centering
	\includegraphics[width=0.96\columnwidth]{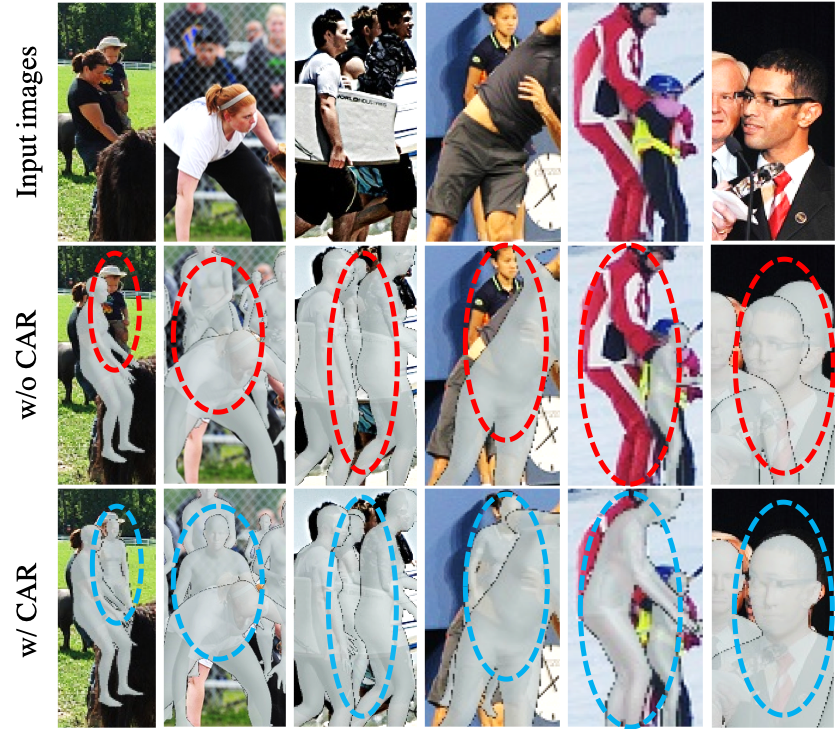}
	\caption{Qualitative ablation study of the CAR on the Crowdpose.}
	\label{fig:CAR_qab}
\end{figure}

\vspace{-1mm}
As shown in Tab.~\ref{tab:3DPW-SPLIT} and \ref{tab:Crowdpose}, CAR improve the PMPJPE metric on the 3DPW-PC and the Crowdpose datasets by 4.8\% and 10.3\%, respectively.
Additionally, Fig.~\ref{fig:CAR_qab} shows, qualitatively, the impact of ablating CAR.
Adding CAR improves  performance in  crowded scenes, which demonstrates that CAR effectively alleviates the center collision problem.

\textbf{Intensity coefficient $\gamma$ of the CAR. }
To set $\gamma$, we conduct an ablation study on 3DPW-PC.
In Tab.~\ref{tab:3DPW-SPLIT}, setting $\gamma=0.2$ performs much better on the crowded scenes (3DPW-PC) and its performance on the normal cases (3DPW-NC/3DPW-OC) is comparable to the best. 
For general in-the-wild cases of Crowdpose, setting $\gamma=0.2$ improves the performance by 10\%  over $\gamma=0$ in Tab.~\ref{tab:Crowdpose}.
Therefore, we suggest training the model with $\gamma=0.2$ for all cases.
The reason performance degrades in the normal cases is probably that pushing apart the body centers affects the consistency of the body-center-guided representation.

\vspace{-1mm}
\subsection{Discussion}
\vspace{-1mm}
To understand the source of our performance gains, we conduct an ablation study on different subsets of 3DPW subsets. 
Results in Tab.~\ref{tab:3DPW-SPLIT} show that our main gains come from the person-occluded and the non-occluded/truncated cases.
It demonstrates the effectiveness of the proposed pixel-level representation in improving the disambiguation in the crowded scenes.
Our experiments suggest that the difference between ROMP and the state-of-the-arts~\cite{jiang2020coherent,kocabas2020vibe,kolotouros2019spin} is the method of representation learning. 
ROMP learns the pixel-level representation from a holistic view, while the multi-stage methods learn a bounding-box-level representation in a local view.
Our one-stage framework enables ROMP to learn more discriminative features that are robust to rich disturbances outside the bounding box, helping generalization. 

\vspace{-1mm}
\section{Conclusion}
\vspace{-1mm}
We introduce a novel one-stage network, ROMP, for monocular multi-person 3D mesh regression. 
For pixel-level estimation, we propose an explicit body-center-guided representation and further develop it as a collision-aware version, CAR, enabling robust prediction under person-person occlusion. 
ROMP is the first open-source one-stage method that achieves state-of-the-art performance on multiple benchmarks as well as real-time inference speed.
ROMP can serve as a simple yet effective foundation for related multi-person 3D tasks, such as depth estimation, tracking, and interaction modeling. 

\noindent \textbf{Acknowledgements:} This work was supported by the National Key R\&D Program of China under Grand No. 2020AAA0103800. 
We thank Peng Cheng for discussing the Center map training.

\noindent 
\noindent\textbf{Disclosure:} MJB has received research funds
from Adobe, Intel, Nvidia, Facebook, and Amazon. While MJB
is a part-time employee of Amazon, his research was performed
solely at Max Planck. MJB has financial interests in Amazon, Datagen Technologies, and Meshcapade
GmbH.

{\small
\bibliographystyle{packages/ieee_fullname}
\bibliography{arxiv}

\begin{thebibliography}{10}\itemsep=-1pt

\bibitem{PoseTrack}
Mykhaylo Andriluka, Umar Iqbal, Eldar Insafutdinov, Leonid Pishchulin, Anton
  Milan, Juergen Gall, and Bernt Schiele.
\newblock Pose{T}rack: {A} benchmark for human pose estimation and tracking.
\newblock In {\em CVPR}, 2018.

\bibitem{mpii}
Mykhaylo Andriluka, Leonid Pishchulin, Peter Gehler, and Bernt Schiele.
\newblock 2{D} human pose estimation: New benchmark and state of the art
  analysis.
\newblock In {\em CVPR}, 2014.

\bibitem{arnab2019exploiting}
Anurag Arnab, Carl Doersch, and Andrew Zisserman.
\newblock Exploiting temporal context for 3{D} human pose estimation in the
  wild.
\newblock In {\em CVPR}, 2019.

\bibitem{bao2020pose}
Qian Bao, Wu Liu, Yuhao Cheng, Boyan Zhou, and Tao Mei.
\newblock Pose-guided tracking-by-detection: Robust multi-person pose tracking.
\newblock {\em IEEE Transactions on Multimedia}, 23:161--175, 2020.

\bibitem{PandaNet_Benzine_2020_CVPR}
Abdallah Benzine, Florian Chabot, Bertrand Luvison, Quoc~Cuong Pham, and
  Catherine Achard.
\newblock Panda{N}et: Anchor-based single-shot multi-person 3{D} pose
  estimation.
\newblock In {\em CVPR}, 2020.

\bibitem{keep}
Federica Bogo, Angjoo Kanazawa, Christoph Lassner, Peter Gehler, Javier Romero,
  and Michael~J Black.
\newblock Keep it {SMPL}: Automatic estimation of 3{D} human pose and shape
  from a single image.
\newblock In {\em ECCV}, 2016.

\bibitem{openpose}
Zhe Cao, Tomas Simon, Shih-En Wei, and Yaser Sheikh.
\newblock Realtime multi-person 2{D} pose estimation using part affinity
  fields.
\newblock In {\em CVPR}, 2017.

\bibitem{cheng2020bottom}
Bowen Cheng, Bin Xiao, Jingdong Wang, Honghui Shi, Thomas~S. Huang, and Lei
  Zhang.
\newblock Higher{HRN}et: Scale-aware representation learning for bottom-up
  human pose estimation.
\newblock In {\em CVPR}, 2020.

\bibitem{choi2020pose2mesh}
Hongsuk Choi, Gyeongsik Moon, and Kyoung~Mu Lee.
\newblock Pose2mesh: Graph convolutional network for 3d human pose and mesh
  recovery from a 2d human pose.
\newblock In {\em ECCV}, 2020.

\bibitem{duan2019centernet}
Kaiwen Duan, Song Bai, Lingxi Xie, Honggang Qi, Qingming Huang, and Qi Tian.
\newblock Centernet: Keypoint triplets for object detection.
\newblock In {\em ICCV}, 2019.

\bibitem{Guler_2019_CVPR}
Riza~Alp Guler and Iasonas Kokkinos.
\newblock Holo{P}ose: Holistic 3{D} human reconstruction in-the-wild.
\newblock In {\em CVPR}, 2019.

\bibitem{resnet}
Kaiming He, Xiangyu Zhang, Shaoqing Ren, and Jian Sun.
\newblock Deep residual learning for image recognition.
\newblock In {\em CVPR}, 2016.

\bibitem{imry2020challengespin}
Kissos Imry, Fritz Lior, Goldman Matan, Meir Omer, Oks Eduard, and Kliger Mark.
\newblock Beyond weak perspective for monocular 3{D} human pose estimation.
\newblock In {\em ECCVW}, 2020.

\bibitem{h36m}
Catalin Ionescu, Dragos Papava, Vlad Olaru, and Cristian Sminchisescu.
\newblock Human3.6{M}: Large scale datasets and predictive methods for 3{D}
  human sensing in natural environments.
\newblock {\em TPAMI}, 2014.

\bibitem{jiang2020coherent}
Wen Jiang, Nikos Kolotouros, Georgios Pavlakos, Xiaowei Zhou, and Kostas
  Daniilidis.
\newblock Coherent reconstruction of multiple humans from a single image.
\newblock In {\em CVPR}, 2020.

\bibitem{lsp}
Sam Johnson and Mark Everingham.
\newblock Clustered pose and nonlinear appearance models for human pose
  estimation.
\newblock In {\em BMVC}, 2010.

\bibitem{lsp_extended}
Sam Johnson and Mark Everingham.
\newblock Learning effective human pose estimation from inaccurate annotation.
\newblock In {\em CVPR}, 2011.

\bibitem{cmu_panoptic}
Hanbyul Joo, Hao Liu, Lei Tan, Lin Gui, Bart Nabbe, Iain Matthews, Takeo
  Kanade, Shohei Nobuhara, and Yaser Sheikh.
\newblock Panoptic studio: A massively multiview system for social motion
  capture.
\newblock In {\em ICCV}, 2015.

\bibitem{joo2020eft}
Hanbyul Joo, Natalia Neverova, and Andrea Vedaldi.
\newblock Exemplar fine-tuning for 3{D} human pose fitting towards in-the-wild
  3{D} human pose estimation.
\newblock In {\em ECCV}, 2020.

\bibitem{hmr}
Angjoo Kanazawa, Michael~J. Black, David~W. Jacobs, and Jitendra Malik.
\newblock End-to-end recovery of human shape and pose.
\newblock In {\em CVPR}, 2018.

\bibitem{kanazawa2019learning}
Angjoo Kanazawa, Jason Zhang, Panna Felsen, and Jitendra Malik.
\newblock Learning 3{D} human dynamics from video.
\newblock In {\em CVPR}, 2019.

\bibitem{kocabas2020vibe}
Muhammed Kocabas, Nikos Athanasiou, and Michael~J Black.
\newblock {VIBE}: Video inference for human body pose and shape estimation.
\newblock In {\em CVPR}, 2020.

\bibitem{kolotouros2019spin}
Nikos Kolotouros, Georgios Pavlakos, Michael~J Black, and Kostas Daniilidis.
\newblock Learning to reconstruct 3{D} human pose and shape via model-fitting
  in the loop.
\newblock In {\em ICCV}, 2019.

\bibitem{gcmr}
Nikos Kolotouros, Georgios Pavlakos, and Kostas Daniilidis.
\newblock Convolutional mesh regression for single-image human shape
  reconstruction.
\newblock In {\em CVPR}, 2019.

\bibitem{kundu2020appearance}
Jogendra~Nath Kundu, Mugalodi Rakesh, Varun Jampani, Rahul~Mysore Venkatesh,
  and R~Venkatesh Babu.
\newblock Appearance consensus driven self-supervised human mesh recovery.
\newblock In {\em ECCV}, 2020.

\bibitem{unite}
Christoph Lassner, Javier Romero, Martin Kiefel, Federica Bogo, Michael~J
  Black, and Peter~V Gehler.
\newblock Unite the people: Closing the loop between 3{D} and 2{D} human
  representations.
\newblock In {\em CVPR}, 2017.

\bibitem{law2018cornernet}
Hei Law and Jia Deng.
\newblock Cornernet: Detecting objects as paired keypoints.
\newblock In {\em ECCV}, 2018.

\bibitem{crowdpose}
Jiefeng Li, Can Wang, Hao Zhu, Yihuan Mao, Hao-Shu Fang, and Cewu Lu.
\newblock Crowd{P}ose: Efficient crowded scenes pose estimation and a new
  benchmark.
\newblock In {\em CVPR}, 2019.

\bibitem{lin2017focal}
Tsung-Yi Lin, Priya Goyal, Ross Girshick, Kaiming He, and Piotr Doll{\'a}r.
\newblock Focal loss for dense object detection.
\newblock In {\em ICCV}, 2017.

\bibitem{coco}
Tsung-Yi Lin, Michael Maire, Serge Belongie, James Hays, Pietro Perona, Deva
  Ramanan, Piotr Doll{\'a}r, and C~Lawrence Zitnick.
\newblock Microsoft coco: Common objects in context.
\newblock In {\em ECCV}, 2014.

\bibitem{liu2018coordconv}
Rosanne Liu, Joel Lehman, Piero Molino, Felipe Petroski~Such, Eric Frank, Alex
  Sergeev, and Jason Yosinski.
\newblock An intriguing failing of convolutional neural networks and the
  coordconv solution.
\newblock In {\em NeurIPS}, 2018.

\bibitem{liu2021recent}
Wu Liu, Qian Bao, Yu Sun, and Tao Mei.
\newblock Recent advances in monocular 2d and 3d human pose estimation: A deep
  learning perspective.
\newblock {\em arXiv}, 2021.

\bibitem{SMPL}
Matthew Loper, Naureen Mahmood, Javier Romero, Gerard Pons-Moll, and Michael~J.
  Black.
\newblock {SMPL}: A skinned multi-person linear model.
\newblock {\em TOG}, 2015.

\bibitem{mono-3dhp2017}
Dushyant Mehta, Helge Rhodin, Dan Casas, Pascal Fua, Oleksandr Sotnychenko,
  Weipeng Xu, and Christian Theobalt.
\newblock Monocular 3{D} human pose estimation in the wild using improved cnn
  supervision.
\newblock In {\em 3DV}, 2017.

\bibitem{mehta2018single}
Dushyant Mehta, Oleksandr Sotnychenko, Franziska Mueller, Weipeng Xu, Srinath
  Sridhar, Gerard Pons-Moll, and Christian Theobalt.
\newblock Single-shot multi-person 3{D} pose estimation from monocular rgb.
\newblock In {\em 3DV}, 2018.

\bibitem{moon2019camera}
Gyeongsik Moon, Ju~Yong Chang, and Kyoung~Mu Lee.
\newblock Camera distance-aware top-down approach for 3{D} multi-person pose
  estimation from a single {RGB} image.
\newblock In {\em CVPR}, 2019.

\bibitem{moon2020i2l}
Gyeongsik Moon and Kyoung~Mu Lee.
\newblock {I2L-MeshNet}: Image-to-lixel prediction network for accurate 3{D}
  human pose and mesh estimation from a single {RGB} image.
\newblock In {\em ECCV}, 2020.

\bibitem{newell2017associative}
Alejandro Newell, Zhiao Huang, and Jia Deng.
\newblock Associative embedding: End-to-end learning for joint detection and
  grouping.
\newblock In {\em NeurIPS}, 2017.

\bibitem{pavlakos2019texturepose}
Georgios Pavlakos, Nikos Kolotouros, and Kostas Daniilidis.
\newblock {TexturePose}: Supervising human mesh estimation with texture
  consistency.
\newblock In {\em ICCV}, 2019.

\bibitem{pishchulin2016deepcut}
Leonid Pishchulin, Eldar Insafutdinov, Siyu Tang, Bjoern Andres, Mykhaylo
  Andriluka, Peter~V Gehler, and Bernt Schiele.
\newblock Deepcut: Joint subset partition and labeling for multi person pose
  estimation.
\newblock In {\em CVPR}, 2016.

\bibitem{redmon2018yolov3}
Joseph Redmon and Ali Farhadi.
\newblock Yolov3: An incremental improvement.
\newblock {\em arXiv}, 2018.

\bibitem{ren2015fasterrcnn}
Shaoqing Ren, Kaiming He, Ross Girshick, and Jian Sun.
\newblock Faster r-cnn: Towards real-time object detection with region proposal
  networks.
\newblock In {\em NeurIPS}, 2015.

\bibitem{rogez2019lcr}
Gr{\'e}gory Rogez, Philippe Weinzaepfel, and Cordelia Schmid.
\newblock Lcr-net++: Multi-person 2{D} and 3{D} pose detection in natural
  images.
\newblock {\em TPAMI}, 2019.

\bibitem{tripletloss}
Florian Schroff, Dmitry Kalenichenko, and James Philbin.
\newblock Facenet: A unified embedding for face recognition and clustering.
\newblock In {\em CVPR}, 2015.

\bibitem{sofiiuk2019adaptis}
Konstantin Sofiiuk, Olga Barinova, and Anton Konushin.
\newblock Adaptis: Adaptive instance selection network.
\newblock In {\em ICCV}, 2019.

\bibitem{centerhmr}
Yu Sun, Qian Bao, Wu Liu, Yili Fu, and Mei Tao.
\newblock Centerhmr: a bottom-up single-shot method for multi-person 3d mesh
  recovery from a single image.
\newblock In {\em arxiv}, 2021.

\bibitem{sun2019dsd-satn}
Yu Sun, Yun Ye, Wu Liu, Wenpeng Gao, YiLi Fu, and Tao Mei.
\newblock Human mesh recovery from monocular images via a skeleton-disentangled
  representation.
\newblock In {\em ICCV}, 2019.

\bibitem{surreal}
G{\"u}l Varol, Javier Romero, Xavier Martin, Naureen Mahmood, Michael~J. Black,
  Ivan Laptev, and Cordelia Schmid.
\newblock Learning from synthetic humans.
\newblock In {\em CVPR}, 2017.

\bibitem{3dpw}
Timo von Marcard, Roberto Henschel, Michael Black, Bodo Rosenhahn, and Gerard
  Pons-Moll.
\newblock Recovering accurate 3{D} human pose in the wild using imus and a
  moving camera.
\newblock In {\em ECCV}, 2018.

\bibitem{aich}
Jiahong Wu, He Zheng, Bo Zhao, Yixin Li, Baoming Yan, Rui Liang, Wenjia Wang,
  Shipei Zhou, Guosen Lin, Yanwei Fu, et~al.
\newblock Ai challenger: A large-scale dataset for going deeper in image
  understanding.
\newblock {\em arXiv}, 2017.

\bibitem{Xu_2019_ICCV}
Yuanlu Xu, Song-Chun Zhu, and Tony Tung.
\newblock {DenseRaC}: Joint 3{D} pose and shape estimation by dense
  render-and-compare.
\newblock In {\em ICCV}, 2019.

\bibitem{zanfir2018monocular}
Andrei Zanfir, Elisabeta Marinoiu, and Cristian Sminchisescu.
\newblock Monocular 3{D} pose and shape estimation of multiple people in
  natural scenes-the importance of multiple scene constraints.
\newblock In {\em CVPR}, 2018.

\bibitem{zanfir2018deep}
Andrei Zanfir, Elisabeta Marinoiu, Mihai Zanfir, Alin-Ionut Popa, and Cristian
  Sminchisescu.
\newblock Deep network for the integrated 3{D} sensing of multiple people in
  natural images.
\newblock In {\em NeurIPS}, 2018.

\bibitem{zhang2021bmp}
Jianfeng Zhang, Dongdong Yu, Jun~Hao Liew, Xuecheng Nie, and Jiashi Feng.
\newblock Body meshes as points.
\newblock In {\em CVPR}, 2021.

\bibitem{zhang2020object}
Tianshu Zhang, Buzhen Huang, and Yangang Wang.
\newblock Object-occluded human shape and pose estimation from a single color
  image.
\newblock In {\em CVPR}, 2020.

\bibitem{zhen2020smap}
Jianan Zhen, Qi Fang, Jiaming Sun, Wentao Liu, Wei Jiang, Hujun Bao, and
  Xiaowei Zhou.
\newblock {SMAP}: Single-shot multi-person absolute 3{D} pose estimation.
\newblock In {\em ECCV}, 2020.

\bibitem{zhou2019objects}
Xingyi Zhou, Dequan Wang, and Philipp Kr{\"a}henb{\"u}hl.
\newblock Objects as points.
\newblock {\em arXiv}, 2019.

\bibitem{ExtremeNet}
Xingyi Zhou, Jiacheng Zhuo, and Philipp Krahenbuhl.
\newblock Bottom-up object detection by grouping extreme and center points.
\newblock In {\em CVPR}, 2019.

\bibitem{Zhou_2019_CVPR}
Yi Zhou, Connelly Barnes, Lu Jingwan, Yang Jimei, and Li Hao.
\newblock On the continuity of rotation representations in neural networks.
\newblock In {\em CVPR}, 2019.

\bibitem{Zhu_2019_CVPR}
Hao Zhu, Xinxin Zuo, Sen Wang, Xun Cao, and Ruigang Yang.
\newblock Detailed human shape estimation from a single image by hierarchical
  mesh deformation.
\newblock In {\em CVPR}, 2019.

\end{thebibliography}
}

\clearpage


\section*{Supplementary material}

# Monocular, One-stage, Regression of Multiple 3D People
## **Supplementary Material**

## 1. Introduction

This supplementary material first provides details about the network architecture design in Sec. 2, followed by hyper-parameter configurations and implementation details in Sec. 3. Finally, we present more qualitative results in Sec. 4, including an analysis of failure cases.

## 2. Ablation study on the architecture

The common architecture underlying most existing methods uses a ResNet-50 backbone followed by a single head that produces SMPL parameters. Most existing methods use an iterative approach, first introduced by HMR [9]. To arrive at the ROMP architecture in Fig. 1, we progressively add/change elements of this basic design and evaluate the effects in an ablation study. As shown in Fig. 2, two main design choices of the head architecture are explored, single head (SH) vs. separated multiple heads (MH), and the number of convolution blocks (NB) in each head. In this section, we first introduce the experiment settings of the ablation study and then report the results.

### 2.1. Experiment settings

**Architecture details.** 1) SH v.s. MH: The architecture of the single head model is presented as (b) in Fig. 2. It is a straightforward and lightweight design, which uses a single branch to jointly estimate the Camera and SMPL maps. In contrast, the separated multi-head design is adopted in Fig. 1. 2) NB: We adopt ResNet blocks (shown in Fig. 1) as the basic unit of the head networks. We conduct an ablation study to determine the proper number of blocks in each head. The architecture is presented as (c) in Fig. 2.

**Datasets.** For pretraining, we take three 2D pose datasets, COCO [15], CrowdPose [14], AICH [20] and one detection dataset, CrowdHuman [18]. For the formal training, we take four 3D pose datasets, Human3.6M [3], MPI-INF-3DHP [16], MuCo-3DHP [16], UP [13], and three 2D pose datasets, COCO [15], MPII [1], LSP [5]. Especially, we only use the 7 subjects (Subject 1, 2, 3, 4, 5, 6, and 7) of MPI-INF-3DHP for training. Subject 8 is used for validation. The test set of MPI-INF-3DHP is employed for evaluation.

| Components | | | | MPI-INF-3DHP | |
|---|---|---|---|---|---|
| SH | MH | NB | Backbone | MPJPE↓ | PMPJPE↓ |
| ✓ |   | 1 | HRNet-32 | 95.31 | 66.73 |
|   | ✓ | 1 | HRNet-32 | **95.11** | 65.89 |
| ✓ |   | 2 | HRNet-32 | 97.19 | **65.50** |
|   | ✓ | 2 | HRNet-32 | 95.30 | 66.32 |
|   | ✓ | 3 | HRNet-32 | 97.19 | 65.94 |
| ✓ |   | 1 | ResNet-50 | 105.31 | 69.62 |
|   | ✓ | 1 | ResNet-50 | 106.43 | 69.83 |
| ✓ |   | 2 | ResNet-50 | 105.37 | **69.12** |
|   | ✓ | 2 | ResNet-50 | **104.92** | 70.20 |
|   | ✓ | 3 | ResNet-50 | 105.41 | 71.78 |

Table 1. Ablation study on the architecture design. SH denotes the single head design and MH is the separated multi-head design. NB is the number of blocks used in each head.

**Training.** Firstly, we pretrain the ResNet-50/HRNet-32 model on the pretraining 2D pose datasets for 120 epochs. The network architecture for pretraining is presented as (a) in Fig. 2. We follow the hyper-parameter settings of Higher-HRNet [2] during this process.

During formal training, we set the learning rate to $5e^{-5}$, weight decay to $1e^{-6}$, batch size to 64. To achieve the best performance of each architecture as much as possible, the loss weights are adjusted, according to the visualization results on the validation set, to avoid the model falling into a local minimum. Each model has been trained at least 100 epochs. We observe that the model with fewer head layers often achieves its best performance at about 120 epochs after several rounds of adjusting hyperparameters, while the model with more head layers often achieves its best performance at about 50 epochs.

### 2.2. Result analysis

In Tab. 1, we present the results of different architecture designs. The experiments are performed with two kinds of backbone, HRNet-32 and ResNet-50. In this settings, HRNet-32 performs better than ResNet-50. Just like its superior performance of fine-tunning on 3DPW (Tab. 3 in the main paper), HRNet-32 once again proves its excellent ability to fit a specific data domain on MPI-INF-3DHP.

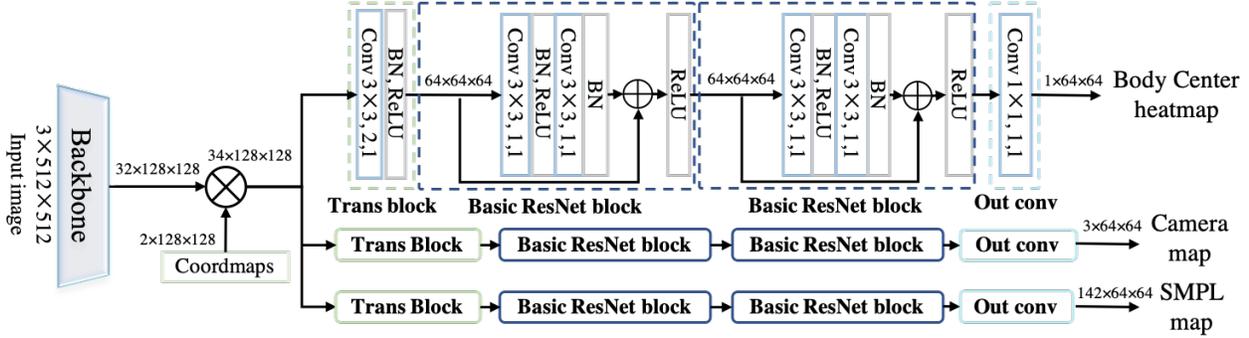

Figure 1. Architecture of the proposed ROMP.

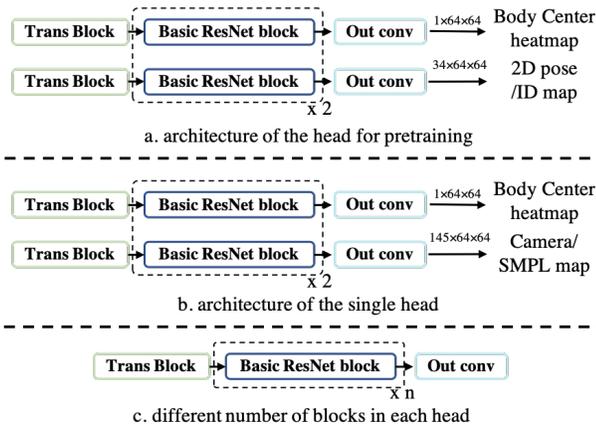

Figure 2. Architecture of the pre-train model.

Using the same backbone, the performance gap between different head architecture designs is relatively small. Among these designs, we find out that 1) compared with the SH, the disentangled multi-head design performs better in the most cases and is prone to train; 2) regarding the NB, setting $n = 2$ is a better choice to balance accuracy and training time.

Fig. 1 shows the details of the architecture. ROMP adopts a fully convolutional multi-head design. Compared with previous methods (like [9, 11, 12]), we do not need to use iterative regression. It is interesting because previous methods rely on this iterative updating to achieve good pose accuracy. ROMP has a harder task than these previous methods that are given a cropped image of the person. ROMP needs to detect people, sort out what image features belong to which person, and estimate the pose. By learning from a holistic view of the whole image, ROMP is forced to learn more about people and how they appear in images. For example, people at the edge of the image usually tend to be truncated. In addition, the holistic view provides the opportunity of learning the interaction between multiple people, which helps handle the crowded scenes. Since ROMP has to solve the pose-estimation problem given an image of the whole scene, it must learn more distinguishable features to solve the task. We posit that these more powerful features enable it to estimate the body shape and pose without iteration.

## 3. Implementation Details

### 3.1. Training Strategy

The basic settings of pretraining and formal training are introduced in Sec. 2.1. Here, we introduce the detailed strategy we used for training ROMP. Our training uses 4 NVIDIA P40 GPUs with a batch size of 128. We adopt the Adam optimizer [10] for training. To avoid the multi-step training and adapt to people of diverse scales, we take both the cropped single-person images and the whole images as input. The ratio of loading the entire images is first set to 10% in the first 60 epochs, and then adjusted to 60% in the remaining 60 epochs. Especially, to accelerate the training and reduce the GPU memory usage, we use the automatic mixed precision (AMP) training of Pytorch [17].

### 3.2. Effect of hyper-parameter configurations during inference

ROMP has several hyper-parameters that can be adjusted during inference to adapt to different scenes, particularly the confidence threshold $t_c$ of the Body Center map and the maximum number of people in an image, $N$. The confidence threshold $t_c$ is used to filter out the detected people with the confidence value lower than $t_c$. Similarly, we set the max person number $N$ to take the top $N$ detected people (sorted by their confidence value on the Body Center map). Changing these parameters only affects the number of output bodies and does not require retraining the model.

We observe that a higher $t_c$ filters out inaccurate/untrusted predictions, while a lower $t_c$ leaves more detection results. The setting of the max person number $N$ follows the same rule. The qualitative ablation study in Fig. 3 illustrates this conclusion. For evaluation on all benchmarks, we have set $t_c = 0.2$ and $N = 64$ for a fair comparison.

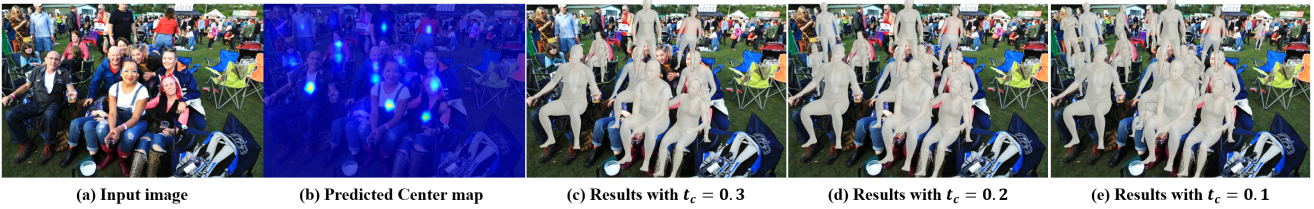

(a) Input image  (b) Predicted Center map  (c) Results with $t_c = 0.3$  (d) Results with $t_c = 0.2$  (e) Results with $t_c = 0.1$

Figure 3. Qualitative ablation study of the confidence threshold $t_c$, on a crowded image.

| Sequence Name | Frame Ranges |
|---|---|
| downtown_bus_00 | 1620-1900 |
| courtyard_hug_00 | 100-500 |
| courtyard_dancing_00 | 60-370 |
| courtyard_dancing_01 | 60-270 |
| courtyard_basketball_00 | 200-280 |
| courtyard_captureSelfies_00 | 500-600 |

Table 2. Video sequences of the person-occluded subset, 3DPW-PC, in 3DPW.

### 3.3. Depth Ordering

For better visualization, we attempt to approximate the depth ordering between the estimated multi-person body meshes to render the meshes onto the original 2D images. In detail, we construct a depth ordering map to determine the visible meshes in front, using 2D body scale and center confidence as the cue. First, we sort by the body center confidence value from largest to smallest. Second, for each body mesh, we compute the 2D area of the body projected onto the image; this gives an approximate measure of its scale. Finally, we adjust the visible mesh at each position according to their 2D areas. Specifically, at a certain position, if the area value of the invisible body mesh is greater than the currently visible body mesh by a threshold, we swap their positions. In this way, we bring to the front, the body mesh that occupies a larger area on the image plane. This is an approximate solution and future work should explore an integrated solution for depth ordering during inference.

### 3.4. Datasets

**3DPW-PC and 3DPW-OC** are the subsets of the 3DPW [19] dataset that are used to evaluate the performance under person or object occlusion respectively. **3DPW-NC** is simply the rest images in the 3DPW. 3DPW-PC contains 1314 frames of 6 person-occluded video sequences. They contain severe person-person occlusion cases with at least two-people overlapping. Details are provided in Tab. 2. Following Zhang et al. [21], 3DPW-OC contains 23 object-occluded video sequences. Please refer to [21] for the details.

**Crowdpose** [14] is a crowded dataset with 2D pose annotations. It contains an abundant variety of person- person and person-object occlusion. Specifically, it contains 20,000 images with about 80,000 persons. We employ their default splits with 9,963 samples for training, 7,991 test samples, and 1,997 validation samples.

**3DOH50K** [21] is a 3D human occlusion dataset. In the image, the human body is occluded by various objects, such as a laptop computer, box, chair, etc. It contains 50,310 images for training and 1,290 images for testing. It is used to evaluate the performance under object occlusion.

**MPI-INF-3DHP** [16] is a single-person multi-view 3D pose dataset. It contains 8 actors performing 8 activities. Over 1.3M frames are captured from all 14 cameras. Except for the indoor RGB videos of a single person, they also provide MATLAB code to generate a multi-person dataset, MuCo-3DHP, via mixing up segmented foreground human appearance.

**COCO** [15], **MPII** [1], **LSP** [5], **LSP Extended** [6], **and AICH** [20] are in-the-wild 2D pose datasets. We use them for training. Especailly, we use the pseudo SMPL annotations of part images generated by [8, 12] for training.

### 3.5. Evaluation Metrics

**Per-vertex error (PVE)** measures the average Euclidean distance from the 3D body mesh predictions to the ground truth after aligning the pelvis keypoints in millimeters.

**The mean per joint position error (MPJPE)** measures the average Euclidean distance from the 3D pose predictions to the ground truth in millimeters. The predictions are first translated to match the ground truth. Generally, they are aligned by the pelvis keypoint for comparison.

**Procrustes-aligned MPJPE (PMPJPE)** is MPJPE after rigid alignment of the predicted pose with ground truth in millimeters. It is also called as the reconstruction error. Through Procrustes alignment, the effects of translation, rotation, and scale are eliminated, thus PMPJPE focuses on evaluating the accuracy of the reconstructed 3D skeleton.

**3D PCK & AUC.** 3D PCK is the 3D version of the 2D PCK metric, which computes the Percentage of Correct Keypoints. Following the ECCV 2020 3DPW Challenge, the threshold of successful prediction is set to 50mm. Correspondingly, the AUC, which is the total area under the PCK-threshold curve, is calculated by computing PCKs by varying the threshold from 0 to 200mm.

**The mean per joint angle error (MPJAE)** measures

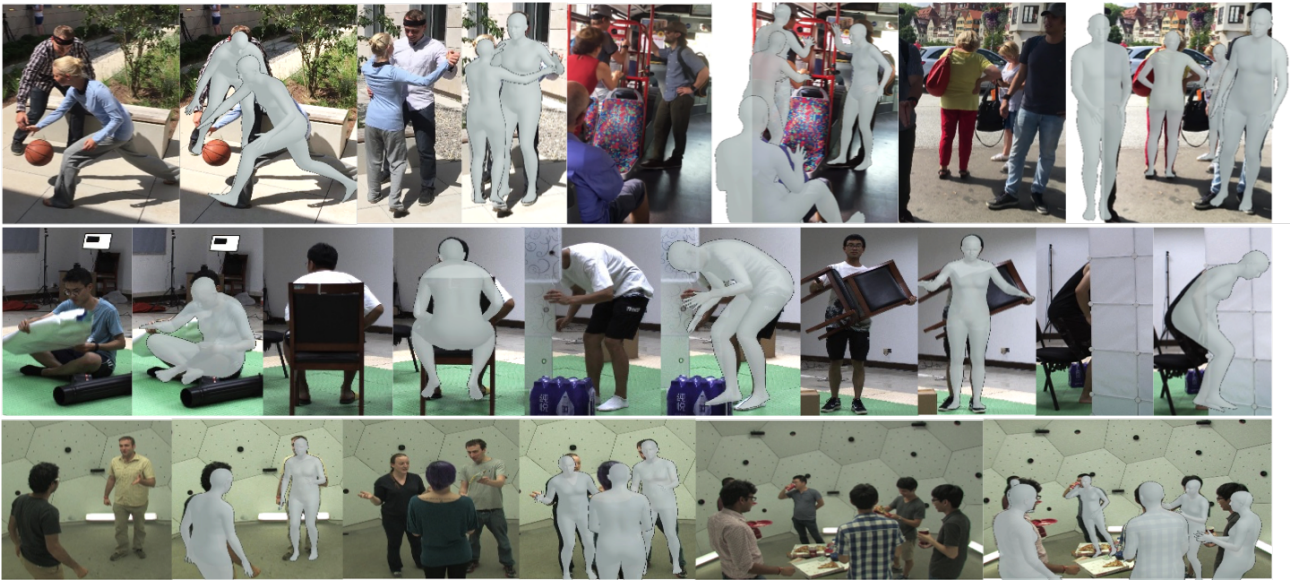

Figure 4. Qualitative results on 3DPW [19], 3DOH50K [21], and CMU Panoptic [7] from top to down.

the angle between the predicted joint orientation and the ground truth orientation in degrees. The orientation difference is measured as the geodesic distance in SO(3). Specifically, only the angles of four limbs and the root are used for evaluation.

**Procrustes-aligned MPJAE (PMPJAE)** measures the MPJAE after applying the rotation matrix, obtained from the Procrustes alignment, on all predicted orientations. As above, it neglects the global mismatch.

**Average Precision (AP)** measures multi-person 2D pose accuracy. We employ it to measure the accuracy of the back-projected 2D body keypoints for evaluating the performance on crowded scenes. A detected keypoint candidate is considered to be correct (true positive) if it lies within a threshold of the ground-truth. Each keypoint separately calculates its correspondence with the ground-truth poses. AP correctly penalizes both missed detections and false positives. In [15], for multi-person pose estimation, AP is further designed by defining the object keypoint similarity (OKS) that is a similarity measure between the predictions and the ground truth. Analog to IoU in object detection, OKS is defined as

$$\text{OKS} = \frac{\sum_i \exp(-d_i^2/2s^2 k_i^2)\delta(v_i > 0)}{\sum_i \delta(v_i > 0)}, \quad (1)$$

where $d_i$ is the Euclidean distance between the detected keypoint and the corresponding ground truth, $v_i$ is the ground-truth visibility flag, $s$ is the person scale, and $k_i$ is a per-keypoint constant that controls falloff. For each keypoint the OKS ranges between 0 and 1.

Given the OKS over all labeled keypoints, average precision (AP) and average recall (AR) can be computed. By tuning OKS values, the precision-recall curve can be calculated, and AP and AR at different OKS can throughly reflect the performance of the testing algorithms. Here, we adopt $\text{AP}^{0.5}$ (AP at OKS = 0.50) and $\text{AR}^{0.5}$ for evaluation.

## 4. Qualitative Results

First, in Fig. 4, we present some qualitative results on evaluation benchmarks that are representative of our results. Next, we present more results on in-the-wild images in Fig. 5. Finally, in Fig. 6, we present some failure cases in estimating depth ordering, detection, and 3D pose. Additionally, we also present results of CRMH [4] on these failure cases for comparison.

**Discussion of failure cases.** Fig. 6 shows the performance of ROMP in estimating the depth order for complex scenes of overlapping people. It illustrates that our body-level depth ordering is limited in cases of extreme crowding with complex depth relationships. CRMH [4] is a Faster-RCNN-based multi-person state-of-the-art method that supervises mesh interpenetration and depth ordering at the vertex level. It has advantages for determining the multi-person depth ordering in crowded scenes. In contrast, ROMP produces more robust and accurate pose estimation in crowded scenes. In future work, we intend to develop a fine-grained depth estimation approach to tackle this problem.

Fig. 6 also shows some extremely challenging images in terms of pose and occlusion. Images like them clearly challenge the state of the art but may also be challenging for humans to perceive.

## References

[1] Mykhaylo Andriluka, Leonid Pishchulin, Peter Gehler, and Bernt Schiele. 2D human pose estimation: New benchmark and state of the art analysis. In *CVPR*, 2014. 1, 3

[2] Bowen Cheng, Bin Xiao, Jingdong Wang, Honghui Shi, Thomas S Huang, and Lei Zhang. Higherhrnet: Scale-aware representation learning for bottom-up human pose estimation. In *Proceedings of the IEEE/CVF Conference on Computer Vision and Pattern Recognition*, pages 5386–5395, 2020. 1

[3] Catalin Ionescu, Dragos Papava, Vlad Olaru, and Cristian Sminchisescu. Human3.6M: Large scale datasets and predictive methods for 3D human sensing in natural environments. *TPAMI*, 2014. 1

[4] Wen Jiang, Nikos Kolotouros, Georgios Pavlakos, Xiaowei Zhou, and Kostas Daniilidis. Coherent reconstruction of multiple humans from a single image. In *CVPR*, 2020. 4

[5] Sam Johnson and Mark Everingham. Clustered pose and nonlinear appearance models for human pose estimation. In *BMVC*, 2010. 1, 3

[6] Sam Johnson and Mark Everingham. Learning effective human pose estimation from inaccurate annotation. In *CVPR*, 2011. 3

[7] Hanbyul Joo, Hao Liu, Lei Tan, Lin Gui, Bart Nabbe, Iain Matthews, Takeo Kanade, Shohei Nobuhara, and Yaser Sheikh. Panoptic studio: A massively multiview system for social motion capture. In *ICCV*, 2015. 4

[8] Hanbyul Joo, Natalia Neverova, and Andrea Vedaldi. Exemplar fine-tuning for 3D human pose fitting towards in-the-wild 3D human pose estimation. In *ECCV*, 2020. 3

[9] Angjoo Kanazawa, Michael J. Black, David W. Jacobs, and Jitendra Malik. End-to-end recovery of human shape and pose. In *CVPR*, 2018. 1, 2

[10] Diederik P Kingma and Jimmy Ba. Adam: A method for stochastic optimization. *arXiv*, 2014. 2

[11] Muhammed Kocabas, Nikos Athanasiou, and Michael J Black. VIBE: Video inference for human body pose and shape estimation. In *CVPR*, 2020. 2

[12] Nikos Kolotouros, Georgios Pavlakos, Michael J Black, and Kostas Daniilidis. Learning to reconstruct 3D human pose and shape via model-fitting in the loop. In *ICCV*, 2019. 2, 3

[13] Christoph Lassner, Javier Romero, Martin Kiefel, Federica Bogo, Michael J Black, and Peter V Gehler. Unite the people: Closing the loop between 3D and 2D human representations. In *CVPR*, 2017. 1

[14] Jiefeng Li, Can Wang, Hao Zhu, Yihuan Mao, Hao-Shu Fang, and Cewu Lu. CrowdPose: Efficient crowded scenes pose estimation and a new benchmark. In *CVPR*, 2019. 1, 3

[15] Tsung-Yi Lin, Michael Maire, Serge Belongie, James Hays, Pietro Perona, Deva Ramanan, Piotr Dollár, and C Lawrence Zitnick. Microsoft coco: Common objects in context. In *ECCV*, 2014. 1, 3, 4

[16] Dushyant Mehta, Helge Rhodin, Dan Casas, Pascal Fua, Oleksandr Sotnychenko, Weipeng Xu, and Christian Theobalt. Monocular 3D human pose estimation in the wild using improved cnn supervision. In *3DV*, 2017. 1, 3

[17] Adam Paszke, Sam Gross, Francisco Massa, Adam Lerer, James Bradbury, Gregory Chanan, Trevor Killeen, Zeming Lin, Natalia Gimelshein, Luca Antiga, et al. Pytorch: An imperative style, high-performance deep learning library. In *NeurIPS*, 2019. 2

[18] Shuai Shao, Zijian Zhao, Boxun Li, Tete Xiao, Gang Yu, Xiangyu Zhang, and Jian Sun. Crowdhuman: A benchmark for detecting human in a crowd. *arXiv preprint arXiv:1805.00123*, 2018. 1

[19] Timo von Marcard, Roberto Henschel, Michael Black, Bodo Rosenhahn, and Gerard Pons-Moll. Recovering accurate 3D human pose in the wild using imus and a moving camera. In *ECCV*, 2018. 3, 4

[20] Jiahong Wu, He Zheng, Bo Zhao, Yixin Li, Baoming Yan, Rui Liang, Wenjia Wang, Shipei Zhou, Guosen Lin, Yanwei Fu, et al. Ai challenger: A large-scale dataset for going deeper in image understanding. *arXiv*, 2017. 1, 3

[21] Tianshu Zhang, Buzhen Huang, and Yangang Wang. Object-occluded human shape and pose estimation from a single color image. In *CVPR*, 2020. 3, 4

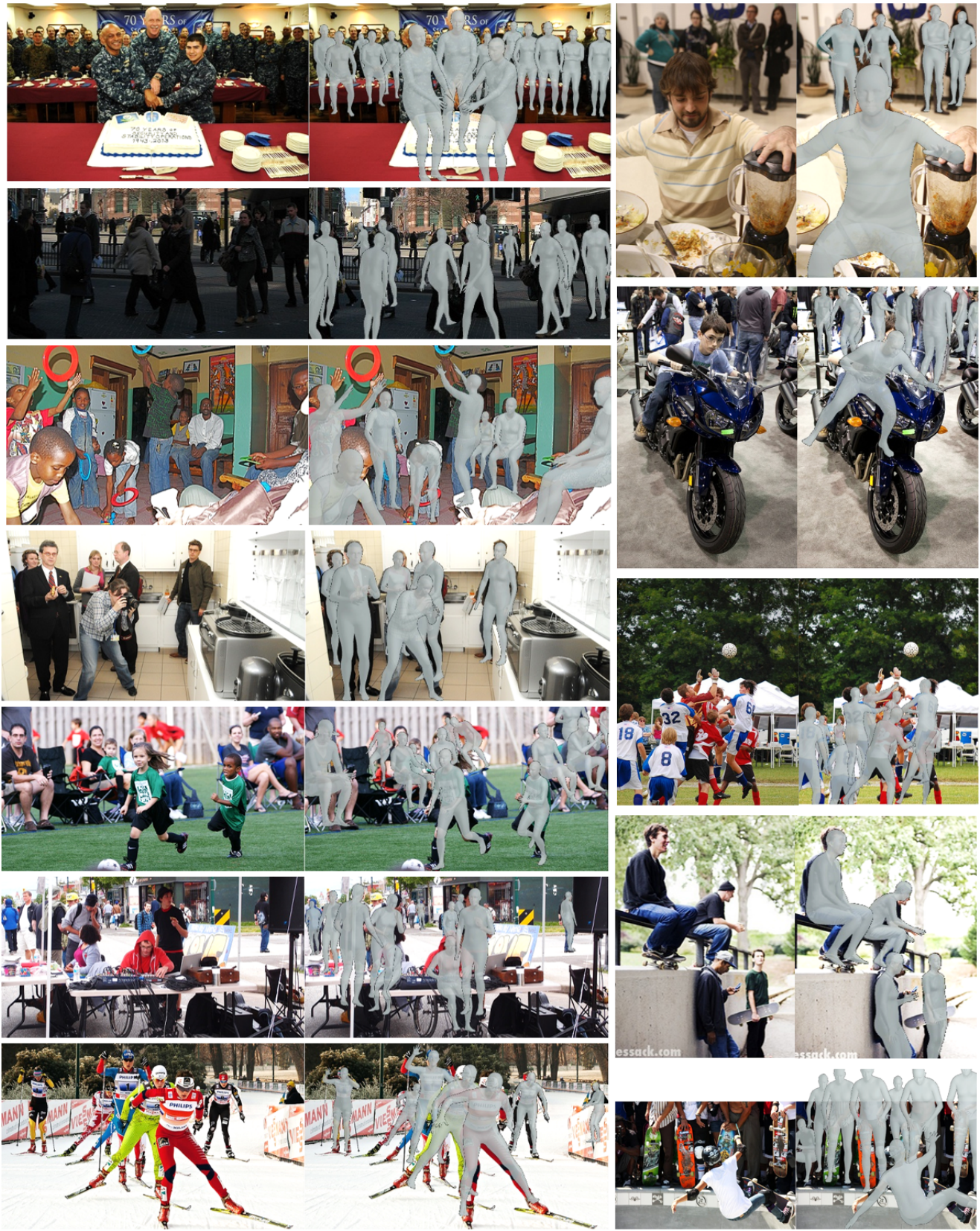

Figure 5. Qualitative results on in-the-wild images.

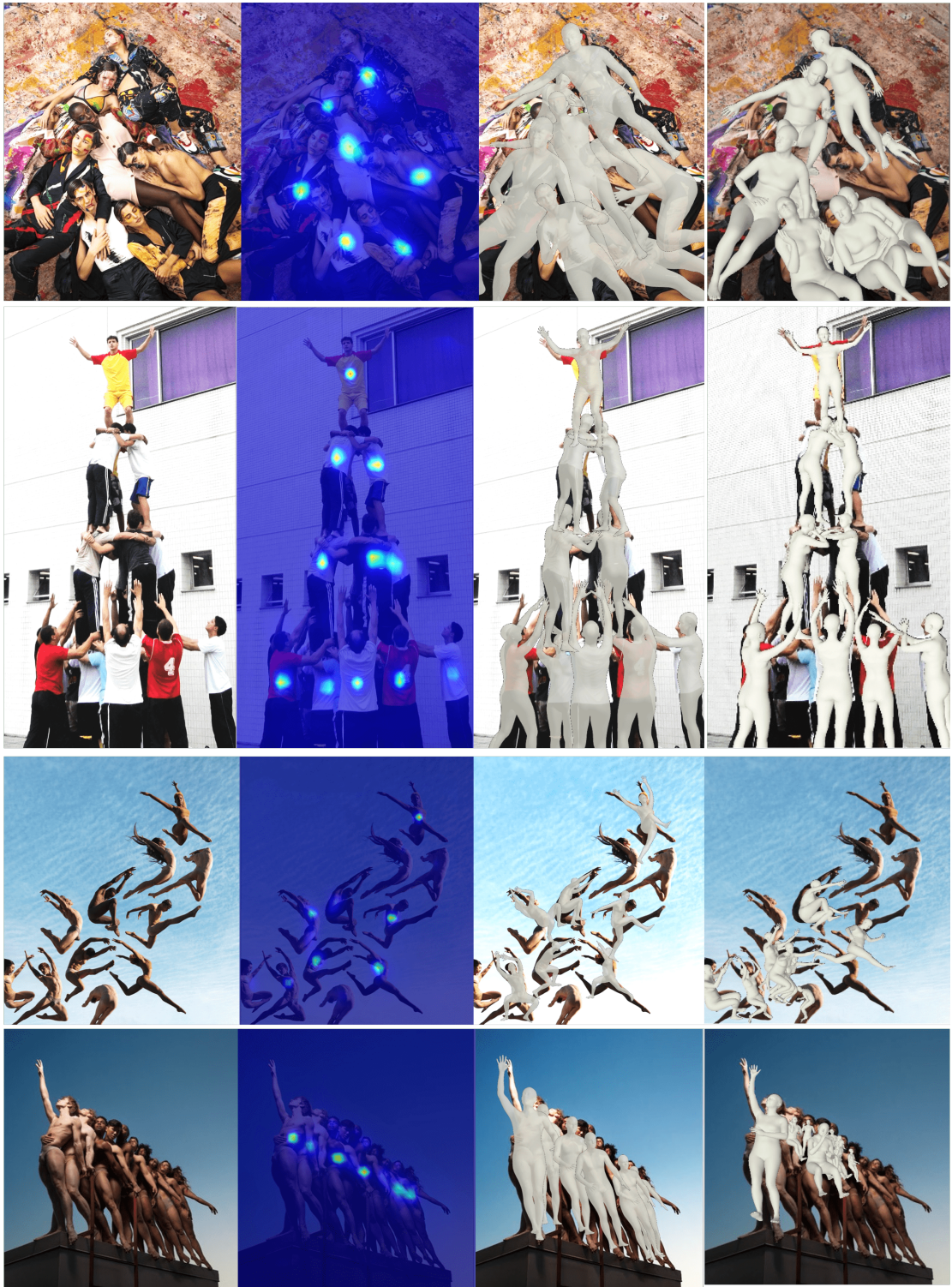

(a) Input images      (b) Predicted Center maps      (c) Our results      (d) Results of CRMH

Figure 6. Failure cases on estimating depth ordering, 3D pose, and detection results in extremely challenging scenes.



\end{document}